\documentclass[11pt,notitlepage,twoside]{report}
\usepackage{url}

\usepackage[T1]{fontenc}
\usepackage[utf8]{inputenc}  
\usepackage{a4wide} 
\usepackage{graphicx} 
\usepackage{caption}
\usepackage{amssymb}
\usepackage{amsmath}
\usepackage{multirow}
\usepackage{listings}
\usepackage[usenames]{color}
\usepackage[left=4cm]{geometry} 

\title{Spectral classification using convolutional neural networks}   
\author{Bc. Pavel Hála} 

\begin{document}
 \definecolor{hellgelb}{rgb}{1,1,0.9	}
 \definecolor{colKeys}{rgb}{0,0,1}
 \definecolor{colIdentifier}{rgb}{0,0,0}
 \definecolor{colComments}{rgb}{1,0,0}
 \definecolor{colString}{rgb}{0,0.5,0}
 \lstset{%
   language={bash},%
    morekeywords={mkdir,paste,cp,ls,rm,mv,xargs,bash,bc,gnuplot}%
 }

 \lstset{%
     float=hbp,%
     basicstyle=\ttfamily\small, %
     identifierstyle=\color{colIdentifier}, %
     keywordstyle=\color{colKeys}, %
     stringstyle=\color{colString}, %
     commentstyle=\color{colComments}, %
     columns=flexible, %
     tabsize=4, %
     frame=single, %
     extendedchars=true, %
     showspaces=false, %
     showstringspaces=false, %
   numbers=left, %
   numberstyle=\tiny, %
   breaklines=true, %
   backgroundcolor=\color{hellgelb}, %
   breakautoindent=true, %
   captionpos=b%
 }

\begin{titlepage}
\begin{center}
\ \\
{\LARGE MASARYK UNIVERSITY}

\Large
{Faculty of Science\\
Department of Theoretical physics and Astrophysics}\\

\vspace{10mm}

\includegraphics[width=4cm]{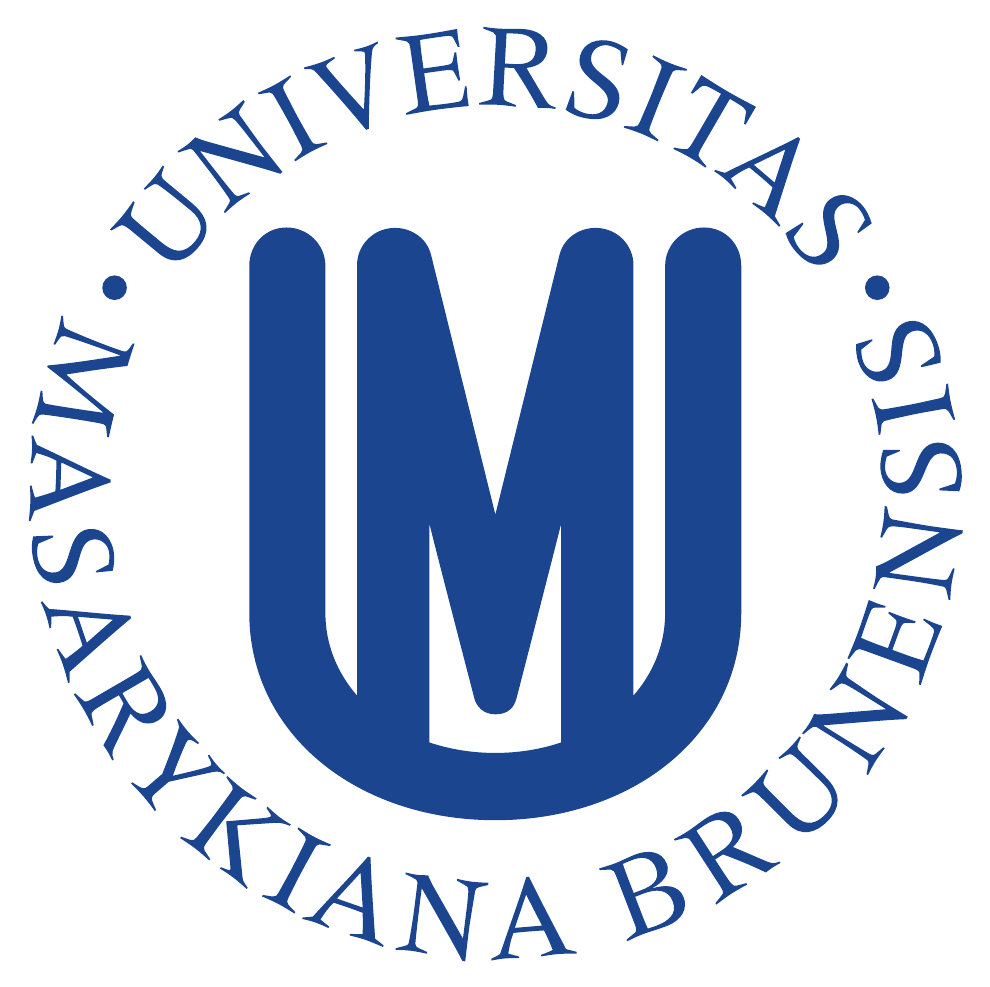} \hspace{3cm} \includegraphics[width=4cm]{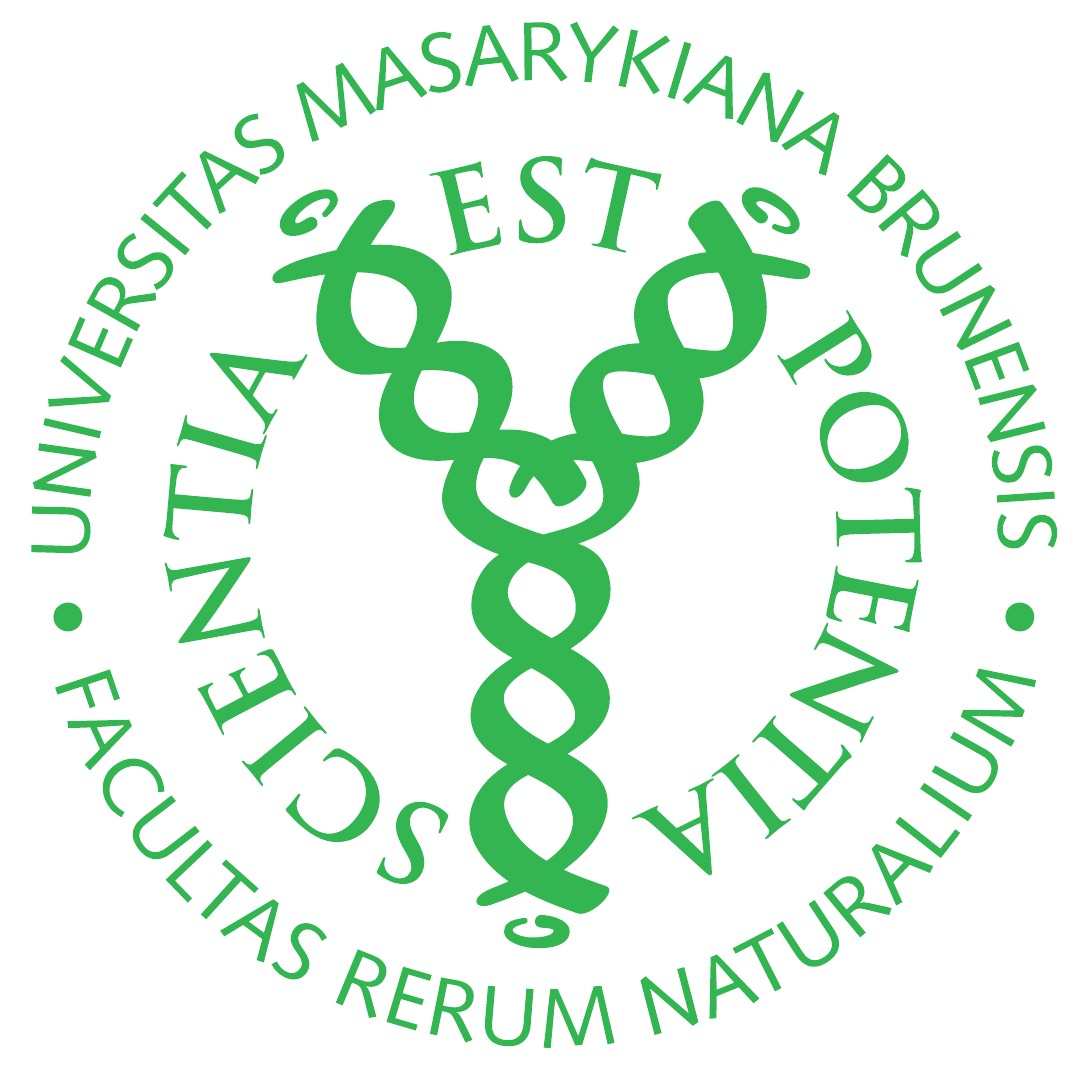}

\vspace{20mm}

{\LARGE\bf Master's thesis}

\vspace{5mm}

{\Large\bf Spectral classification using convolutional neural networks}\\ 

\vspace{20mm}

{\LARGE Bc. Pavel Hála}\\ 

\vspace{40mm}

Brno 2014 
\end{center}
\end{titlepage} 

\pagestyle{empty}
\newpage


\chapter*{Bibliographic entry}

\thispagestyle{empty}

\begin{tabbing}
\hspace{1.75in}\=\kill
\textbf{Author:} \> Bc. Pavel Hála \\
	\> Faculty of Science, Masaryk University \\
	\> Department of Theoretical Physics and Astrophysics \\[1em]
\textbf{Title of thesis:} \> Spectral classification using\\
	\> convolutional neural networks \\[1em]
\textbf{Degree programme:} \> Physics \\[1em]
\textbf{Field of study:} \> Theoretical Physics and Astrophysics \\[1em]
\textbf{Supervisor:} \> Mgr. Filip Hroch, Ph.D. \\[1em]
\textbf{Year of defence:} \> 2014 \\[1em]
\textbf{Keywords:} \> classification, spectra, galaxies,\\
	\> deep learning, artificial intelligence, quasars,\\
	\> stars, convolutional neural network

\end{tabbing}

\newpage

\normalsize 
\noindent

\noindent Abstract: \\

\noindent There is a great need for accurate and autonomous spectral classification methods in astrophysics. This thesis is about training a convolutional neural network (ConvNet) to recognize an object class (quasar, star or galaxy) from one-dimension spectra only. Author developed several scripts and C programs for datasets preparation, preprocessing and postprocessing of the data. EBLearn library (developed by Pierre Sermanet and Yann LeCun) was used to create ConvNets. Application on dataset of more than 60000 spectra yielded success rate of nearly 95\%. This thesis conclusively proved great potential of convolutional neural networks and deep learning methods in astrophysics. \\

\noindent

\newpage

\ \vspace{10mm}

\noindent I'd like to thank my girlfriend Elena Lindišová for her moral support during research that led to writing this thesis.
\bigskip
\bigskip
\bigskip
\bigskip

\noindent This research has made use of the HEAsoft software for astrophysical data analysis, created by NASA Goddard Space Flight Center.
\bigskip

\noindent Funding for SDSS-III has been provided by the Alfred P. Sloan Foundation, the Participating Institutions, the National Science Foundation, and the U.S. Department of Energy Office of Science. The SDSS-III web site is http://www.sdss3.org/.

SDSS-III is managed by the Astrophysical Research Consortium for the Participating Institutions of the SDSS-III Collaboration including the University of Arizona, the Brazilian Participation Group, Brookhaven National Laboratory, Carnegie Mellon University, University of Florida, the French Participation Group, the German Participation Group, Harvard University, the Instituto de Astrofisica de Canarias, the Michigan State/Notre Dame/JINA Participation Group, Johns Hopkins University, Lawrence Berkeley National Laboratory, Max Planck Institute for Astrophysics, Max Planck Institute for Extraterrestrial Physics, New Mexico State University, New York University, Ohio State University, Pennsylvania State University, University of Portsmouth, Princeton University, the Spanish Participation Group, University of Tokyo, University of Utah, Vanderbilt University, University of Virginia, University of Washington, and Yale University.
\bigskip

\vspace{\fill} 
\noindent I declare that I wrote my diploma thesis independently and exclusively using sources cited. I agree with borrowing the work and its publishing.

\bigskip
\bigskip
\bigskip
\noindent In Brno, May 14th 2014\hspace{\fill}Pavel Hála\\ 


\newpage

\pagestyle{plain}
\setcounter{page}{5} 

\tableofcontents 

\newpage 


\chapter{Introduction}

\section{Motivation}
My diploma thesis should be considered as a loose continuation of my bachelor's thesis. In that work \cite{bakal}, I utilised state of the art technology (recently launched Fermi's LAT telescope) to create spectral studies of quasars and blazars with unprecedented angular precision and time resolution. It gave me two important impressions. 

The first was that there is vast amount of spectral event data. Although I studied only several objects, I was dealing with substantial amount of data and it took me a lot of time and effort to obtain even basic spectral charasteristics. Large scale study of hundreds or thousands of objects is unimaginable for an individual and very difficult for a team of people. And the amount of data is rising each day as the Fermi telescope continues in collecting fresh data.

Second impression was that obtained data were very noisy and chaotic. I had to deploy sophisticated algorithms and mathematical procedures in order to reconstruct the spectra. I often wasn't able to find logic behind those algorithms due to their complexity. This feeling wasn't unjustified because many of these algorithms were later updated or rewritten. The telescope was brand new and instrumental functions were afflicted by incorrect preflight assumptions of engineers who developed the instrument. 

Both of these impressions led me to hunt for an alternative approach to spectral analysis of active galaxies. I found out there already is a branch of astrophysics dealing with the problems I mentioned above. It's an emerging field of astroinformatics. I began to experiment with artificial neural networks, namely \texttt{FANN} library\cite{fann}. I spent two years mastering feedforward ANNs, developing optimization algorithms to enhance their sucess rate. I applied them not only in astrophysics, but also in econometrics and quantitative finance. I came to conclusion that although for some applications their performance is pretty good, human time invested into pre and post processing has to be extensive. The results were very sensitive to network parameters and data normalization. This definitely isn't how I imagine an artificial intelligence. Luckily less than a year ago I began to study deep learning, only recently discovered and the most promising field of artificial intelligence. And I was literally blown away by it's performance. I decided this is the path for more efficient analysis of galactic spectra.

When I had the tool I needed to find a source of suitable spectral data. It had to offer diverse and large enough database of specra. Besides, I needed the spectra to be correctly classified with reliable basic spectral parameters. This is not something Fermi's data archive could offer. After some research I decided to proceed with Sloan Digital Sky Survey (SDSS), namely the BOSS \cite{boss}, which is aimed on study of galaxies and quasars. SDSS has a long track record of quality data capture with correct spectral classification. Archive of the data acquired by BOSS instrument contains approximately 1.5 million spectra.

\section{Goals}

This diploma thesis is focused on the algorithmic part of the spectral analysis. The main effort is development of efficient method of spectral classification. I want to highlight that I won't focus on physical interpretation of spectral features of specific objects. Rather I will try to develop a universal framework everybody can easily use for determination of whatever spectral features he wishes. Reliability, adaptability and robustness are the qualities I want to maximize.

My goal is to simulate real conditions of spectral analysis. In that situation astronomers often don't have any information about the source they are observing and are unaware of instrumental discrepancies of the detector. As an input into my algorithm I will use only spectra themselves. No additional information about the source will be included. By studying the data from SDSS archive, I found out there are not many spectral parameters that are reliably defined. Actually only class of the object (galaxy, quasar, star) and its redshift are determined for every object.
\newline
\newline
Goals of the thesis are:
\begin{itemize}
\item Clarify the background of deep learning and introduce the most advanced representative of computer vision --- convolutional neural networks.
\item Use a convolutional neural network to construct an algorithm capable of reliable object type classification (galaxy, quasar, star) based on object's optical spectrum only.
\item Test the algorithm on large subset of spectra and evaluate its accuracy.
\end{itemize}
\noindent

\newpage

These goals will be particularly hard given the nature of the classification. Choosing the correct type of object can be often unambiguous, especially selecting between galaxy and quasar. This is exactly the type of human-like classification that is hard to automate. But the size of the database (1.5 mil objects) underlines the need for robust automation methods. If sucessful, most promising will be utilization of this method in future data-intensive missions without long history of spectral analysis tools like the SDSS.

\newpage ~ \newpage

\chapter{Data description}

\section{SDSS-III DR10}
I've been using data from the third Sloan Digital Sky Survey in its tenth data release (DR10). It's one of the largest sky surveys available and its data are freely available to the public. Major changes in Data Release 10 are the new infrared spectra from APOGEE spectrograph and addition of hundreds of thousands spectra from the BOSS instrument. For my purpose is important that DR10 incorporates new reductions and galaxy parameter fits for all the BOSS data.

\begin{figure}[!htbp]
\begin{center}
\includegraphics[width=14cm]{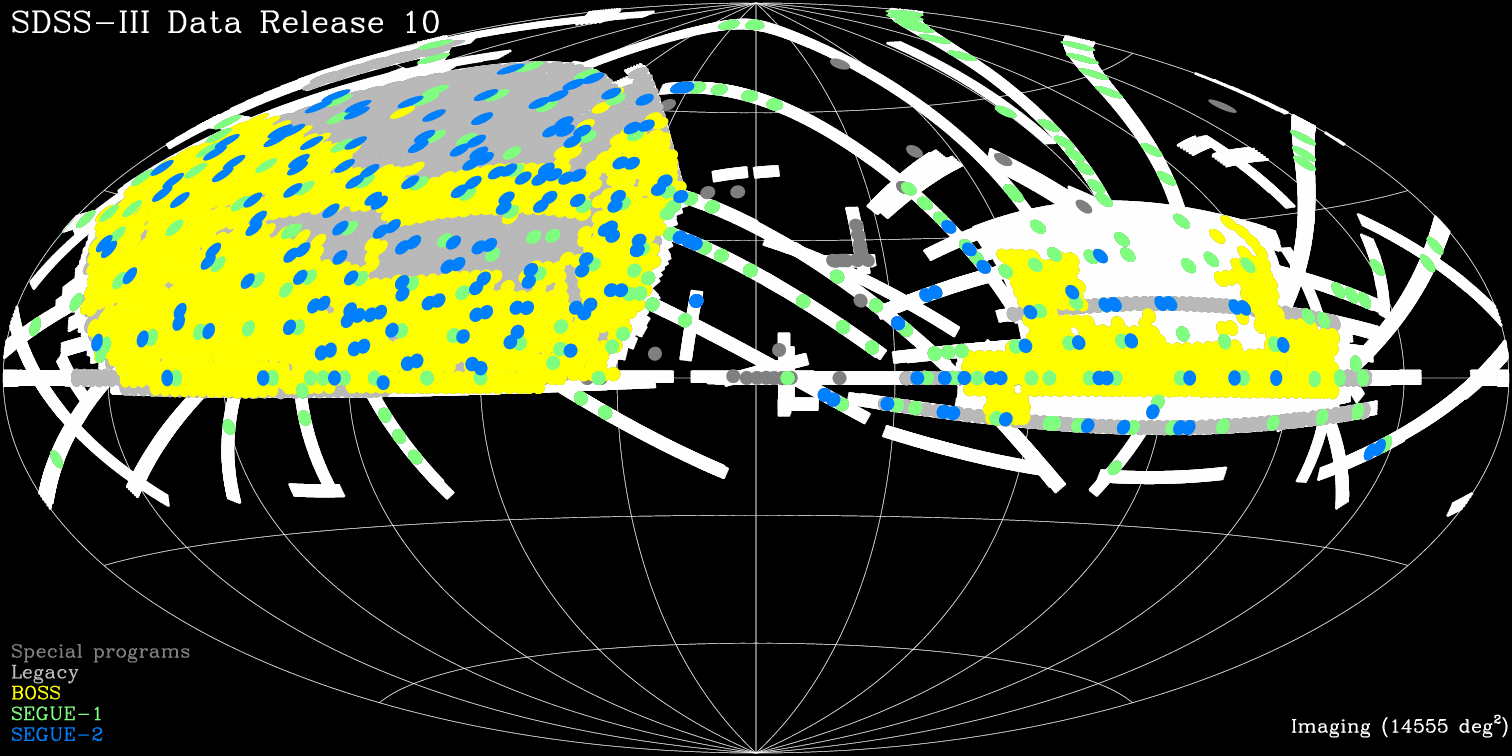}
\caption{DR10 SDSS/BOSS sky coverage \cite{sdss}.}
\label{pic_sdss_coverage}
\end{center}
\end{figure}

\section{BOSS}

BOSS stands for Baryon Oscillation Spectroscopic Survey. It maps the spatial distribution of luminous red galaxies and quasars to detect the characteristic scale imprinted by baryon acoustic oscillations in the early universe. Sound waves that propagated in the early universe, like spreading ripples in a pond, imprinted a characteristic scale on cosmic microwave background fluctuations. These fluctuations have evolved into today's walls and voids of galaxies, meaning this baryon acoustic oscillation scale (about 150 Mpc) is visible among galaxies today.

\begin{figure}[!htbp]
\begin{center}
\includegraphics[width=10cm]{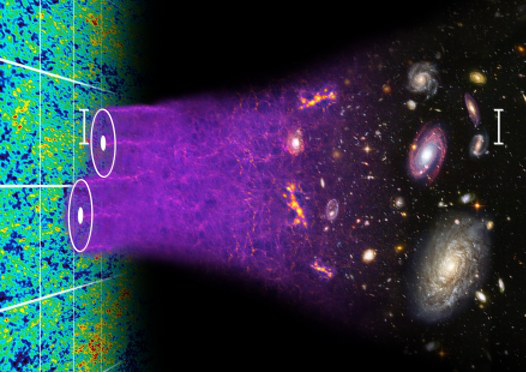}
\caption{Illustration of baryon acoustic oscilations \cite{boss}.}
\label{pic_bao}
\end{center}
\end{figure}

For my purpose it's important that BOSS's main target are galaxies and quasars. In order to map distibution of galaxies and quasars it has to identify all objects first. I assume there is a great emphasis placed on the precision of the identification. That's important. For example the BOSS's own classification pipeline is doing a subclassification after it classifies the object as a galaxy, quasar or star. In this subclassification one estimates stellar class (if the object is a star) or galaxy subtype (starburst, starforming, etc.). Previously, I wanted to include this subclass in my algorithm. But I subsequently learned that this subclass identification in the BOSS catalog is quite inaccurate. It has never been a priority and there is no plan to fix it \cite{boss_caveats}.

\section{Spectroscopic Pipeline}
BOSS project uses \texttt{idlspec2d} software for automated object classification and redshift measurement \cite{pipeline}. The spectral classification and redshift-finding analysis is performed as a $\chi^2$ minimization. Linear fits are made to each observed spectrum using multiple templates and combinations of templates evaluated for all allowed redshifts, and the global minimum-$\chi^2$ solution is adopted for the output classification and redshift. The least-squares minimization is performed by comparison of each spectrum to full range of templates spanning galaxies, quasars, and stars\cite{pipeline}.

\begin{figure}[!htbp]
\begin{center}
\includegraphics[width=10cm]{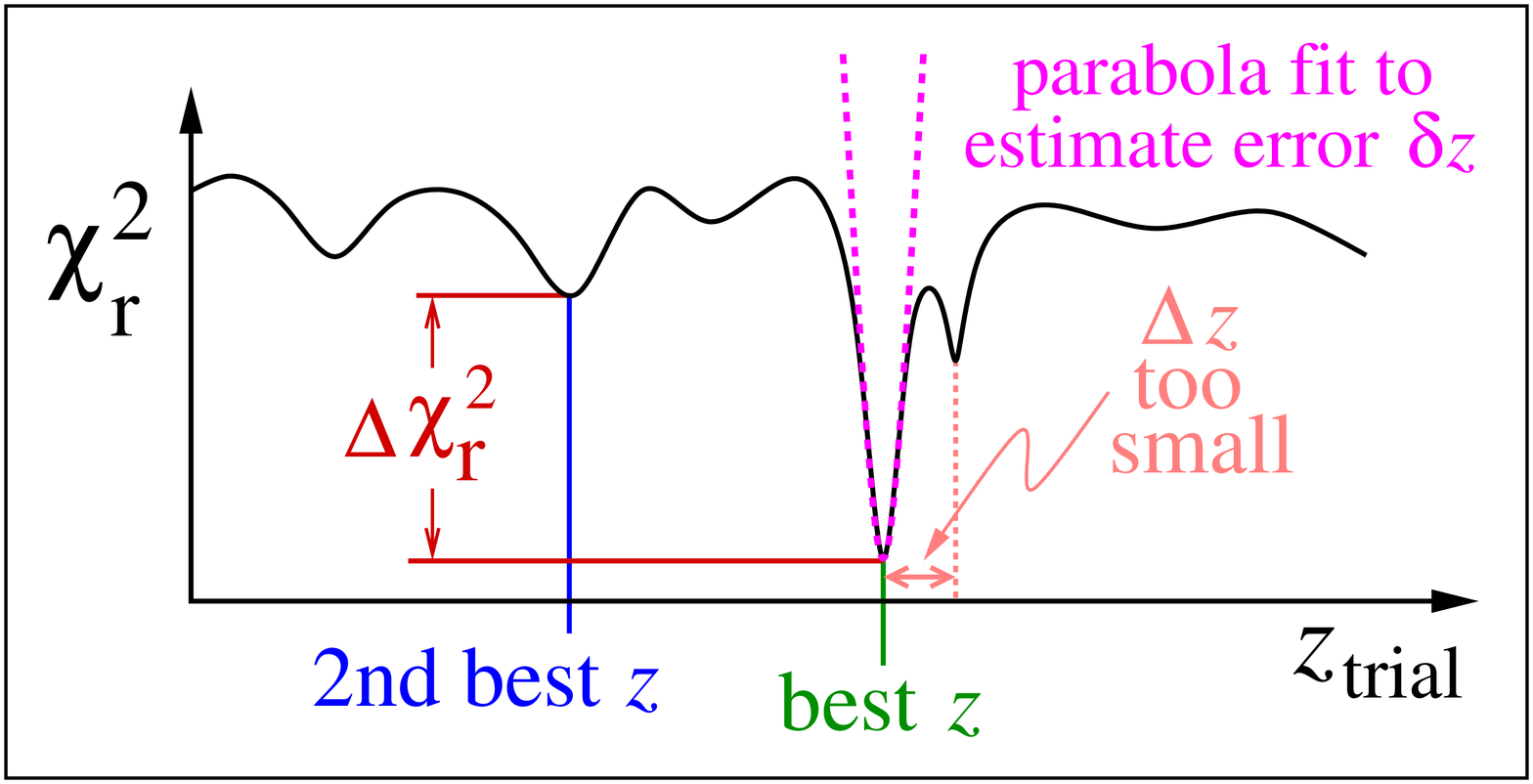}
\caption{Redshift $\chi^2$ minimization process \cite{pipeline}.}
\label{pic_chi2}
\end{center}
\end{figure}

The processed spectra are cut off into 3600\,\AA--10400\,\AA\ range. \texttt{Z\_WARNING} flag is assigned to each spectrum after the classification. Only spectra with \texttt{Z\_WARNING} equal to zero are considered good for scientific analysis. 

\begin{table}[!b]
\begin{tabular}{l|l|p{9cm}}
Bit & Name & Definition \\
\hline
0 & \texttt{SKY} & Sky fiber \\
1 & \texttt{LITTLE\_COVERAGE} & Insufficient wavelength coverage\\
2 & \texttt{SMALL\_DELTA\_CHI2} & $\Delta \chi_r^2$ between best and second-best fit <0.01\\
3 & \texttt{NEGATIVE\_MODEL} & Synthetic spectrum negative\\
4 & \texttt{MANY\_OUTLIERS} & More than 5\% of points above 5$\sigma$ from synth. spectrum\\
5 & \texttt{Z\_FITLIMIT} & $\chi^2$ minimum for best model at the edge of the redshift range\\
6 & \texttt{NEGATIVE\_EMISSION} & Negative emission in a quasar line >3$\sigma$\\
7 & \texttt{UNPLUGGED} & Broken or unplugged fiber\\
\end{tabular}
\caption{Redshift and classification warning flags \cite{pipeline}.}
\label{tab_zwarning}
\end{table}

Overall precision of object classification and redshift is highly dependent on many variables. One of them is the quality of templates used for fitting observed spectra. Another critical parameter is $\delta\chi_{r}^{2}$ in the minimization process (figure \ref{pic_chi2}). Alternative method with less parameters would be desirable. Especially for sky surveys with not as long track record as SDSS.

\section{NOQSO parameter}
I had an issue with galaxy spectra. According to \cite{boss_caveats}, all the galactic spectral parameters have to use the \texttt{NOQSO} appendix. For instance, instead of \texttt{Z} to use \texttt{Z\_NOQSO}, etc. It's because the QSO emission lines can fit galaxy absorption lines resulting in erroneous fits. Using the \texttt{NOQSO} appendix assures the QSO templates are excluded when performing classification/redshift fits. 

However my personal experience is that using \texttt{NOQSO} appendix leads to incorrect object classes and erroneous redshifts. I suspect the information on the SDSS website \cite{boss_caveats} and in the corresponding paper \cite{pipeline} to be relevant for older data releases only. Further clarification would be helpful from somebody in the BOSS team. I made several mass crosschecks and verified random group of spectra manually. That reassured me to don't use \texttt{NOQSO} parameter.

\chapter{Deep learning}

\section{Artificial intelligence}
In the first half of the 20th century, people thought about artificial intelligence (AI) in direct connection with robots. During decades, advancement in robotics proved to be quite disconnected with advancement in AI. Today we have robots all around us, but they tend to be somewhat dumb. Our modern computers have exponentially higher computational performance than our grandparents would imagine. However powerful doesn't meen intelligent. Even the comically looking robots from black-and-white films from the 30s, respond more intelligently than today's hexacore servers. Artificial intelligence we are seeking is the ability to behave efficiently in a real world. That's an environment with blurred inputs and unexpected events. 

\section{Classification based on pattern recognition}
Real development in the field of artificial intelligence started to happen in second half of the 20th century. And ironically up to this date, most promising way seems to be simulation of real biological processes in living creatures. I will be doing spectral classification in this thesis. Therefore I have to describe how an AI-like approach to pattern recognition was originally developed. The following description is based on paper \cite{ann_bouchain} by David Bouchain.

The problem in general can be described as a classification of input data, represented as vectors, into several categories. We need to find a suitable classification function $F$ that fits the equation

\begin{equation}
\mathbf{y}^{p} = F(\mathbf{x}^{p}, W)
\label{ann_classifier}
\end{equation}

\noindent where $p$ is the number of pattern, $\mathbf{x}^{p}$ is the p-th input vector (an image, spectrum, etc), $\mathbf{y}^{p}$ is the output vector, and $W$ is a set of trainable parameters for the classifier (\ref{ann_classifier}). If there exists a vector $\mathbf{T}^{p}$ representing the desired output for each input vector $\mathbf{x}^{p}$, then $\Theta$ is called a training set and is defined as

\begin{equation}
\Theta = \left\lbrace (\mathbf{x}^{p}, \mathbf{T}^{p}) : \mathbf{x}^{p} \in \mathbb{R}^{d}, \mathbf{T}^{p} \in \mathbb{R}^{n}, p = 1,..., P \right\rbrace 
\label{ann_train}
\end{equation}

Adjusting or training the parameters based on comparison between output $\mathbf{y}^{p}$ and label $\mathbf{T}^{p}$ is called a supervised learning. To make a good classifier we have to determine the amount of adjustment for the parameters $W$. This is done through error function $E(W)$ for all patterns, or $E^{p}(W)$ for the p-th pattern and returns error of the classifier. Our task during learning is to minimize this error. Probably the best method for pattern recognition is called gradient-based learning. The idea is to sequentially adjust the parameters $W = W(t)$, where $t$ is the current training step. The minimization of $E^{p}(W)$ will occur by using

\begin{equation}
W(t+1) = W(t) - \epsilon \frac{\partial E^{p}(W(t))}{\partial W(t)}
\label{ann_grad}
\end{equation}

\noindent equation for each pattern. $\epsilon$ is a learning rate.

\section{Multilayer perceptron}

One of the first representatives of this approach were the artificial neural networks (ANNs) \cite{wiki_ann}. It's a computational model inspired by the nervous system. It's capable of learning patterns in the underlying data (inputs) and can be trained to give us anticipated answers (outputs). It can handle noisy data and can cope with highly nonlinear data. It was sucessfully applied to AI-like types of problems (hard or impossible to solve by computers) like handwriting recognition or speech recognition.

\begin{figure}[!htbp]
\begin{center}
\includegraphics[width=10cm]{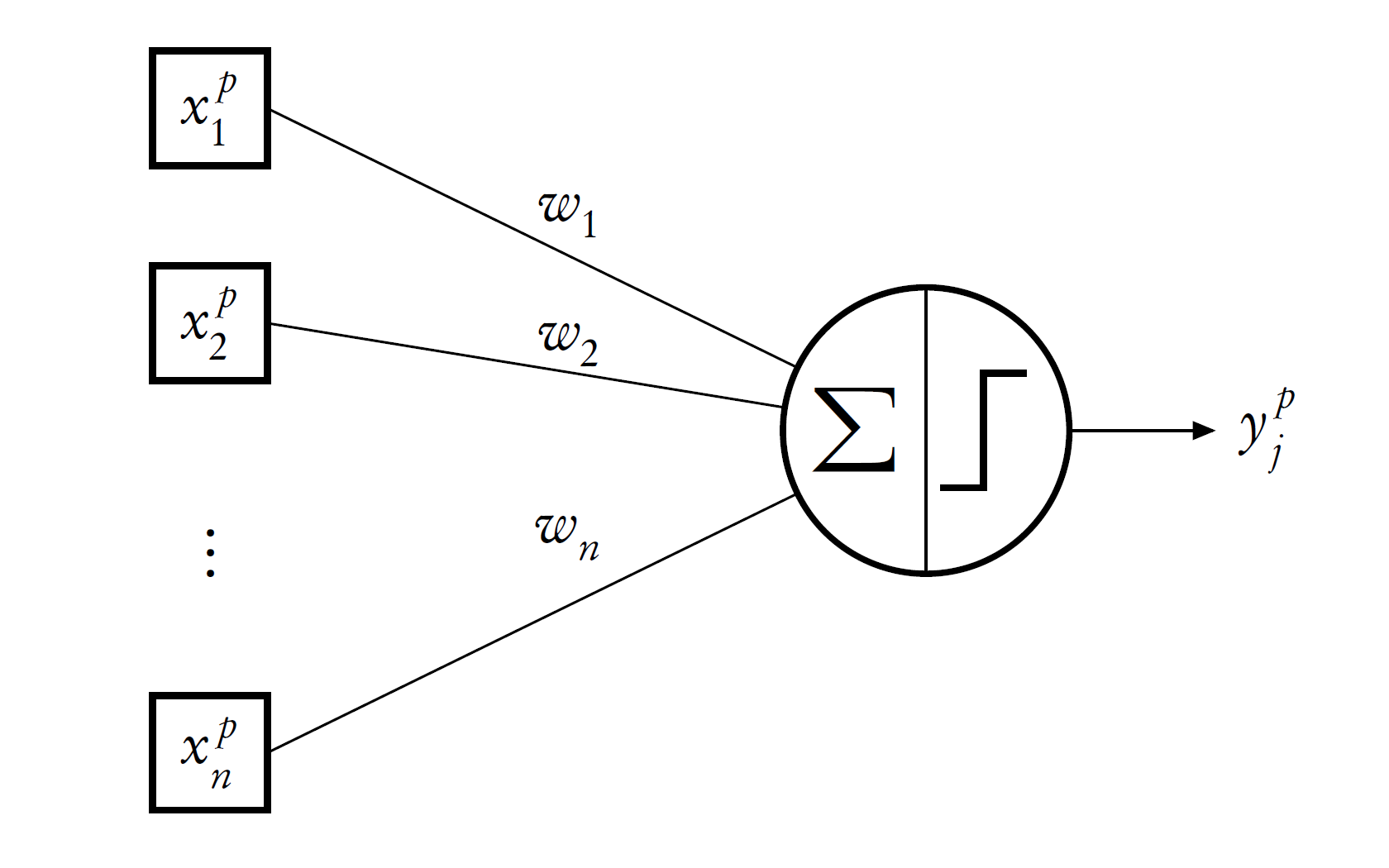}
\caption{Schematics of a perceptron \cite{ann_bouchain}.}
\label{pic_neuron}
\end{center}
\end{figure}

ANN is composed of interconnected artificial neurons. Like a biological neuron has dendrites, artificial neuron has inputs that are feeding the data into the neuron. Each input has assigned weight. Weighted sum is made of all inputs and the result is fed into the activation function. That is continuous, usually (but not necessarily) highly nonlinear function. The neuron amplifies the inputs according to their relevance in minimizing the error rate and transforms the output.

The model of neuron in figure \ref{pic_neuron} is called a perceptron. Inputs from other neurons form a vector $\mathbf{x}$. Each input has weight assigned to it. All inputs from the vector $\mathbf{x}$ are multiplied by corresponding weights from the vector $\mathbf{w}$. The product is summed up and bias $\theta$ is added.

\begin{equation}
u = \sum_{k=1}^{n}\mathbf{x}_{k}\mathbf{w}_{k} - \theta
\label{ann_perceptron}
\end{equation}

Activation function $f(u)$ amplifies/deamplifies it and produces output $y$ from the neuron. Most common acivation function is a sigmoid function in equation (\ref{ann_actfunc}) (others are gauss, stepwise, linear, etc.). The equation (\ref{ann_transform}) and the whole process is obviously similar to the equation (\ref{ann_classifier}).

\begin{equation}
y = f(u)
\label{ann_transform}
\end{equation}

\begin{equation}
f(u) = \frac{1}{1 + e^{-\beta u}}
\label{ann_actfunc}
\end{equation}

\begin{figure}[!b]
\begin{center}
\includegraphics[width=10cm]{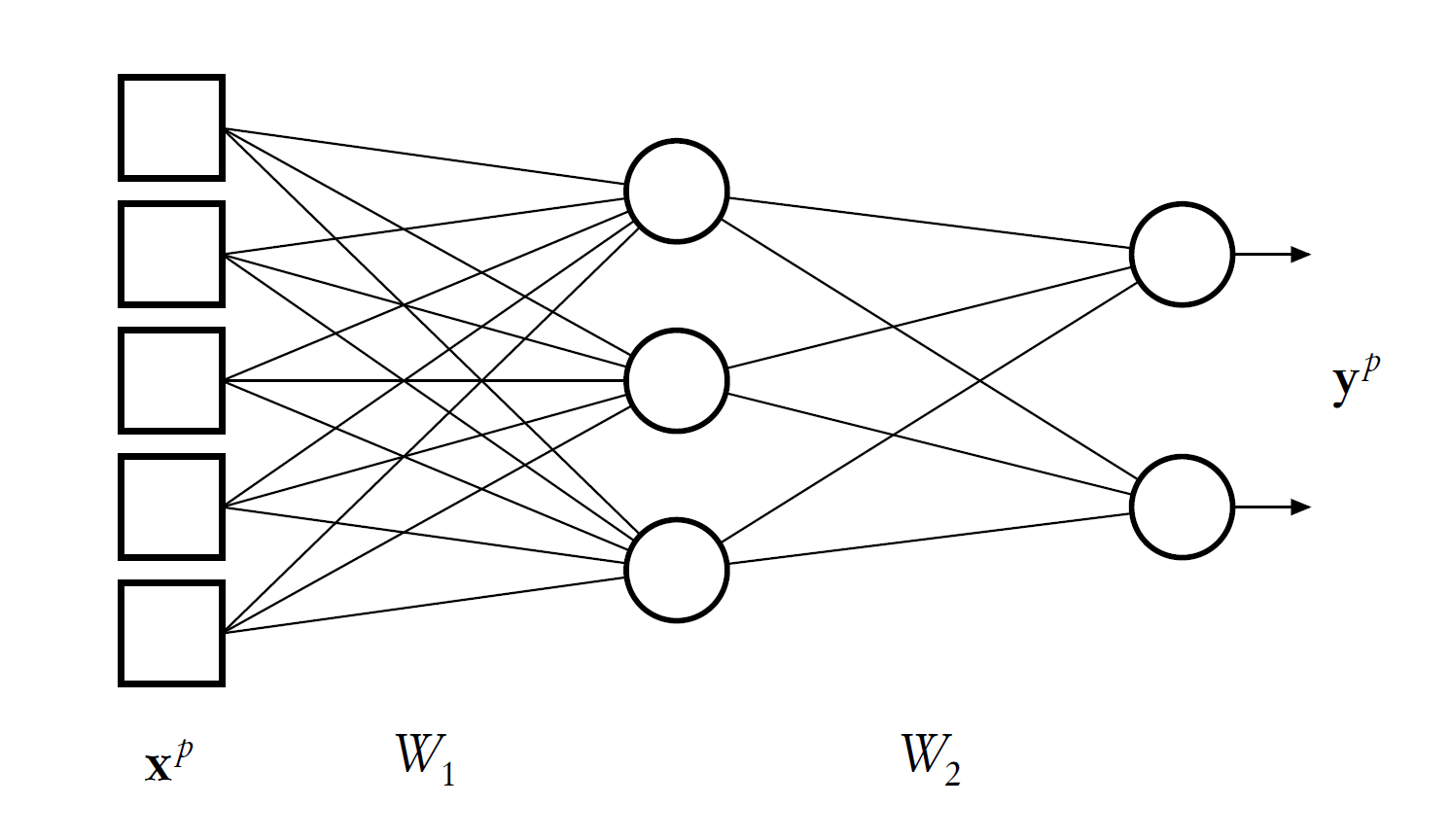}
\caption{Multilayer perceptron model \cite{ann_bouchain}.}
\label{pic_perceptron}
\end{center}
\end{figure}

Multiple perceptrons can be connected together to form a multilayer perceptron (MLP). MLPs are usually fully connected 3-layer networks. First layer consists of input neurons that don't perform any calculation. They just source the data and distribute them to neurons in the second, hidden layer. Hidden layer and output layer follow the steps described in equations (\ref{ann_perceptron}) and (\ref{ann_transform}). MLP is a feedforward neural network model. Information flows only one way from the input layer through the hidden layer to the output layer. MLP uses supervised learning by backpropagation algorithm. Supervised learning means that inputs are labeled. Therefore the network knows what's a desired output from each input dataset. After each run of the network, errors are computed by comparing the output to the desired output. Errors then propagate back through the network and algorithm computes deltas of all weights. Finally all weights are updated and the network is ready for another run.

The error function $E(W)$ over all patterns $p$ is defined as

\begin{equation}
E(W) = \sum_{p}E^{p}(W) = \sum_{p}\left|\left|\mathbf{T}^{p}-\mathbf{y}^{p} \right|\right|^{2}
\label{ann_err}
\end{equation}

\noindent where $\mathbf{y}^{p}$ is output from the output layer, $\mathbf{T}^{p}$ is desired output from equation (\ref{ann_train}). $W$ is a matrix of weights $w_{ij}$ between $i$-th and $j$-th neuron. After differentiation of $E^{p}$ by $w_{ij}$ and slight modifications the learning rule for neurons in the output layer is obtained:

\begin{equation}
w_{ij}(t+1) = w_{ij}(t) + \epsilon y_{i} \delta_{j}
\label{ann_backout}
\end{equation}
\begin{equation}
\delta_{j} = (T_{j} - y_{j}) f'(u_{j})
\end{equation}

\noindent where $y_{j}$ is output of neuron $j$ in the output layer, $y_{i}$ is output of neuron $i$ in the hidden layer and $T_{j}$ is desired output. Learning function for the hidden layer is then

\begin{equation}
w_{ki}(t+1) = w_{ki}(t) + \epsilon x_{k} \delta_{i}
\label{ann_backhid}
\end{equation}
\begin{equation}
\delta_{i} = \sum_{j} \delta_{j} w_{ij} f'(u_{i})
\end{equation}

\noindent where $j$ respresents neurons from output layer, $i$ neurons from hidden layer and $k$ input neurons. $u_{i}$ and $u_{j}$ are respective dendrite potentials.

Important issue in neural nets is bias and variance. It is directly related to overfitting which is a common problem in ANNs. There is a tendency to craft the architecture and parameters of the net in a way that keeps its ability to generalize. The net is always trained only on a subset of available data. Good ANN should be able to perform sucessfully on data it hasn't seen before. We tend to keep the network compact, with only as little number of neurons as possible. Huge net quickly overfits during training. Rather than learn decisive features of the system, it fits to the noise and its performance on out-of-sample data is poor. 

On the other hand if we shrink the network too much, its simplicity prevents it to learn fine patterns in the data. These patterns are often crucial for correct classification. The net performs similarly on all datasets --- similarly poor.

\section{Review of the multilayer perceptron}
I have multiyear experience with artificial neural networks, namely the multilayer perceptron with a single hidden layer. I've been using \texttt{FANN} library written in C \cite{fann}. I've written various utility bash scripts and optimization C programs for this library. I've come to some important conclusions about MLPs that are otherwise difficult to assume. There is a great deal of misconception about neural networks. They are sometimes unjustifiably glorified or being blindly denounced on the other hand. As always, the truth is somewhere in the middle.
\newline
\newline
\noindent Examples of use on a physical system include classification, prediction of future states of the system, finding if and how the system influences other systems or just discovering hidden patterns in the data. According to my experience, these are the \textbf{advantages} of MLPs: 
\begin{itemize}
\item \textbf{Versatility} - MLP can be applied to many fields. From physics to econometrics or biology. And best thing about it is that you don't have to be an expert in those fields to produce results as good as experts. You just have to create good enough training sample and encode the inputs and outputs correctly.
\item \textbf{Multidimensionality} - You made an observation of some phenomenon and need to make a prediction of future states of the system or just how measured physical quantities depend on themselves. When you have two dimensional problem ($y$ somehow depends on $x$), it's easy. You make a chart, do linear regression, polynomial regression, etc. When it's three dimensional, it's still fine. When it's four or five dimensional, it's hard. When it's fifty dimensional, it's nearly impossible. Not so with neural networks. Dimensionality of the input data is just a matter of scalability in case of MLP.
\item \textbf{Scalability} - Neural nets are easily scalable. If you are trying to model physical system that can be described by one or two differential equations, it's quite easy. You create a simple model and you're done. But when the system gets more complicated, your model is useless. You then have to create a completely new and much more complicated model. With neural nets, you only adjust few parameters, add some neurons to the hidden layer and that's it. It takes you few minutes to scale the network, so it's able to model behaviour of exponentially more complicated system.
\item \textbf{Chaotic data} - There are many papers such as \cite{ann_chaos} or \cite{ann_comparison} that demonstrate very good ability of neural networks to work even on very chaotic data. Systems manifesting deterministic chaos are nearly impossible to model with classical methods. That's the reason why ANNs are often used for weather numerical prediction models.
\item \textbf{Human-like} - Probably the greatest feature of neural nets is their ability to model human-like behaviour. That's because human emotions are often hard to predict. In this kind of applications can ANNs prove why they really earn the title of artificial intelligence. To demostrate this, I suggest you to try the simple \texttt{brain.js} demo developed by Heather Arthur \cite{brainjs}. It's a web based app you can use to train a simple JavaScript neural network to recognize your preferred color contrast. After sucessful training you can test it on various colors and compare the results of the network to YIQ formula \cite{wiki_yiq}. General feeling is that the network learns your habits really well. However results will differ from person to person. By testing it on myself I found out that in colors where network's result is conflicting with YIQ formula, I usually prefer the network's version. 
\end{itemize}

\noindent However MLP neural networks have also significant \textbf{disadvantages} such as:
\begin{itemize}
\item \textbf{It's a black box} - You can sucessfully model almost any physical system with ANN. However the ANN source the inputs and produce outputs (no matter how sucessfuly). You don't learn anything about the mechanism or processes inside the physical system. This is undoubtedly a weakness. But this is how we expect an artificial intelligence to work. There are even some problems, often involving classification, that can't be described by an actual mechanism or set of equations.
\item \textbf{Not suitable for everything} - MLP is a universal tool. However that doesn't mean it's always the best option. There are much simpler tools with better performance for many kind of tasks. It's wise to deploy MLP only on problems where it would be beneficial.
\item \textbf{Need for optimization} - Before training the ANN, we must specify some parameters of the network. There are not too many of them, usually much less than in classical models. But in classical models, we usually guess most of these parameters from nature of the problem. In MLP, these parameters are unrelated to the problem. The talk is about net structure, number of neurons in hidden layers, type of activation functions, steepness of the activation functions, number of training epochs, etc. I tested genetic algorithms to estimate these parameters and found out they work, but often drastically increase the training time. I personally didn't use them regularly and prefer the manual estimation. It's quick and with some experience yields the same results as GA. I also tried Cascade training algorithm in \texttt{FANN} library \cite{fann_cascade}, but found it not very useful. The algorithm basicaly tries to find the optimal architecture of the network. Training starts with empty net and neurons are continuously added one by one into the layers until the best configuration is found. The algorithm often diverged and and created absurdly huge nets with poor performance.
\item \textbf{Bias and variance} - There is no best solution to this problem. The line between good generalization and overfitting is very thin. Cross validation on out-of-sample data is necessary. Bias to different datasets is important. Careful preparation and normalization of training datasets is a priority and have huge impact on quality of the result.
\end{itemize}

My work with feedforward neural networks, namely multilayer perceptrons, was aimed on predicting behaviour of various systems. Application on physical systems was perfect. I got precise results and training wasn't complicated. But things get interesting when I tried something more complicated. I wanted to predicit chaotic econometric data. And the results were no longer so good. I was able to get successful prediction only in some periods when network was lucky and identified a pattern in recent data. And even in that case, it identified correctly only the direction of the change, estimating value of the change was impossible. 

In contradiction to widespread beliefs, performance was best when training set was very small, comprised from recent data only. Raising number on neurons in the hidden layer proved to have little effect on success rate. It only added to overfitting. I also found that essential parameters of the network (number of neurons in hidden layer, steepness of activation functions and number of training epochs) strongly depends on each other. For instance increasing number of hidden neurons had the same effect as rising the amount of training epochs. 

Multilayer perceptrons are powerful, versatile and revolutional. They are very capable regressors. But nothing more. I miss the ability to abstract and cope with the problem more independently, without the requirement for massive optimization.

\section{Shallow and deep architectures}
The model of multilayer perceptron I was reviewing in previous section was a shallow architecture. These are architectures with zero or only one hidden layer. Using MLP with only one hidden layer is a common practice. That's because in 1989, George Cybenko proved that a three-layer neural network (MLP with one hidden layer) with a continuous sigmoidal activation functions can approximate all continuous, real-valued functions with any desired precision \cite{deep_cybenko}.

But ever since the birth of artificial neural networks in the 40's, there has been an interest in deep architectures - those with multiple hidden layers. Why's that? There are several reasons:

\begin{itemize}
\item Shallow architectures offer little or no invariance to shifting, scaling, and other forms of distortion.
\item Topology of the input data is completely ignored in shallow networks, yielding similar training results for all permutations of the input vector.
\item Visual system in human brain is deep. It has about ten to twenty layers of neurons from the retina to the inferotemporal cortex (where object categories are encoded).
\end{itemize}

\section{Visual cortex}
\begin{figure}[!htbp]
\begin{center}
\includegraphics[width=10cm]{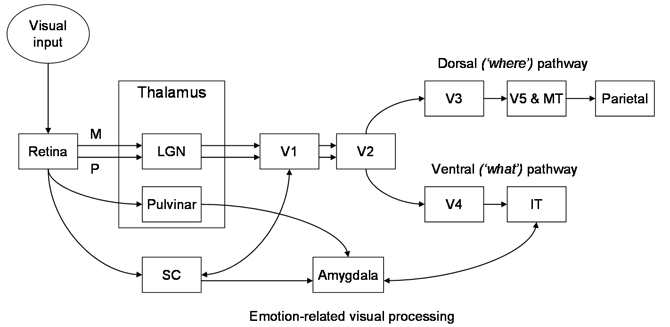}
\caption{Model of the human visual processing system \cite{deep_vis_system}.}
\label{pic_vis_system}
\end{center}
\end{figure}

With advancements in human neuroscience, it was discovered that human visual system is in fact a deep architecture. This had huge impact on the field of computer vision, where feedforward networks such as multilayer perceptron belong. It questioned the wide belief that shallow architectures are the best. It was shown that human ability to abstract requiers human visual cortex to have a deep structure.

\begin{figure}[!htbp]
\begin{center}
\includegraphics[width=3.4cm]{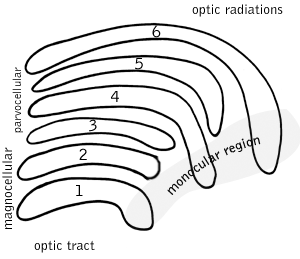}
\caption{Lateral geniculate nucleus (LGN) in the human brain \cite{deep_wiki_lgn}.}
\label{pic_lgn}
\end{center}
\end{figure}

The visual input goes through layers of retina and retinal ganglion cells to lateral geniculate nucleus (LGN) as can be seen in the figure \ref{pic_vis_system}. LGN consists of six layers of neurons. Each layer receives information from the retinal hemi-field of one eye. All the neurons in LGN regions form a topographical map of the visual field from its projection onto the retina \cite{deep_v1}.

\begin{figure}[!htbp]
\begin{center}
\includegraphics[width=10cm]{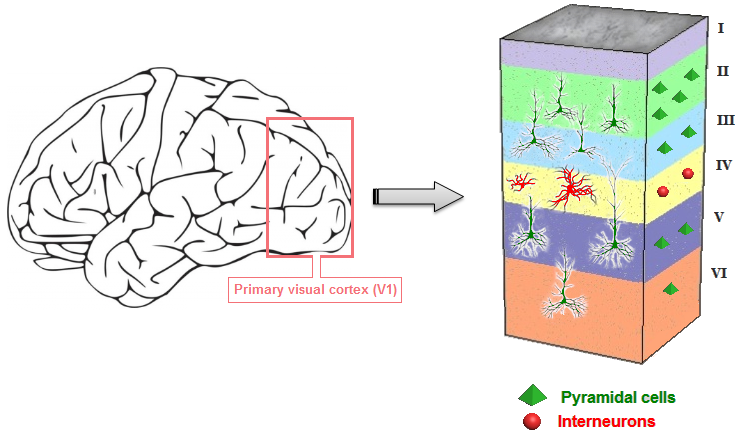}
\caption{Primary visual cortex (V1) \cite{deep_v1}.}
\label{pic_v1}
\end{center}
\end{figure}

From the LGN, the signal goes into the primary visual cortex (V1 in figure \ref{pic_vis_system}). The primary visual cortex is an important and most studied area of the human visual system. It's a six layer, deep neural network with different types of vertically and laterally connected neurons. The figure \ref{pic_v1} shows its structure. Neurons in V1 responds primarily to basic properties of an object, such as edges, length, width or motion \cite{deep_v1}.

\begin{figure}[!htbp]
\begin{center}
\includegraphics[width=5cm]{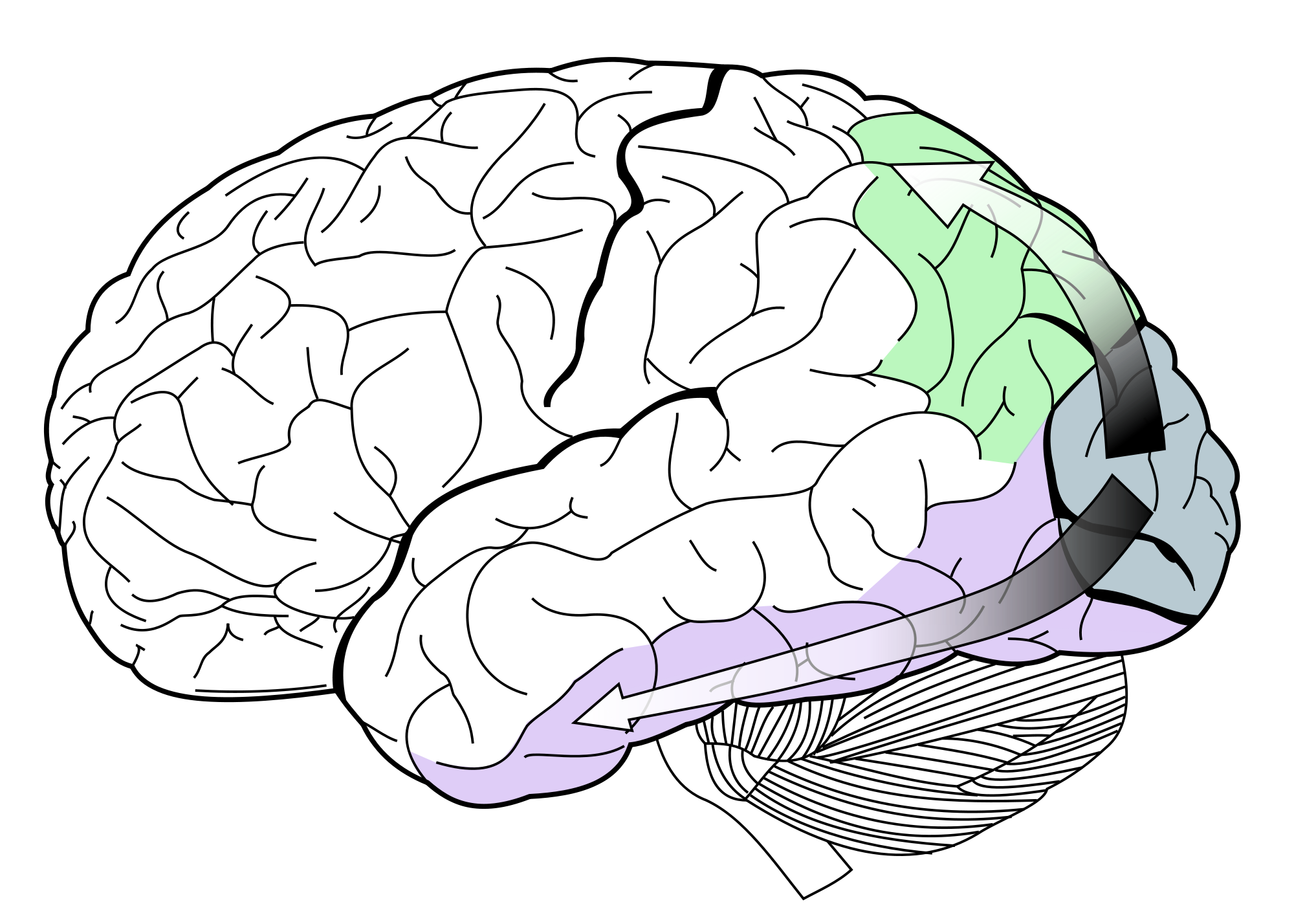}
\caption{Primary visual cortex - V1 (blue area), dorsal pathway (green) and ventral pathway (purple) \cite{deep_wiki_v1}.}
\label{pic_pathways}
\end{center}
\end{figure}

\begin{figure}[!htbp]
\begin{center}
\includegraphics[width=14cm]{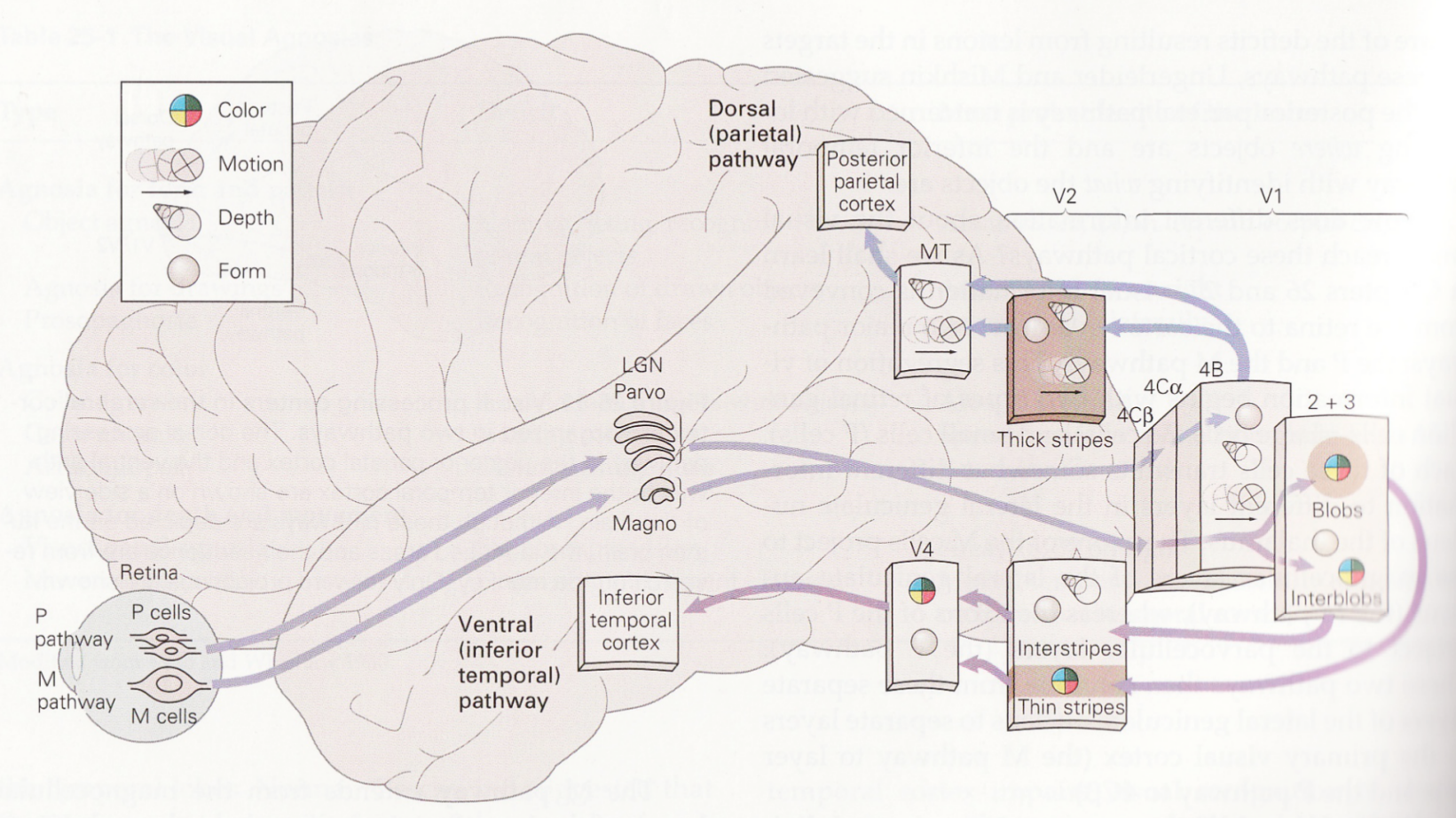}
\caption{Human visual system \cite{deep_present}.}
\label{pic_streams}
\end{center}
\end{figure}

V1 is an important crossroads for the visual signal. From there, the visual signal is transmitted to two streams --- dorsal pathway and ventral pathway. Illustration of this process is in figure \ref{pic_pathways} and \ref{pic_streams}. Dorsal pathway begins in V1 and continues through V2 and V5/MT into the posterior parietal cortex. Is is responsible for "where" characteristics and is associated with motion, representation of object's location and control of the eyes and arms. Ventral pathway also begins in V1. The signal goes through V2, visual area V4 into the inferior temporal cortex. It's called a "what pathway" and is associated with form recognition and object representation \cite{deep_wiki_v1}. Obviously the ventral stream will be my point of interest.

Ventral pathway stands as a model for the computer vision. I has deep architecture and hierarchically abstracts features of an object. Image, which is at the retina level represented as a matrix of pixels, is run through the ventral stream. In V1, edges of the object are detected, V2 detects primitive shapes and in V4 higher level features and colors are abstracted. Inferior temporal cortex produces "what" the object is, i.e. assigns a label and category to the object.

\section{How to build a deep network}

Theoretically you can create a deep neural network by adding hidden layers to the multilayer perceptron and making the layers two dimensional (we need to process images). However there are few problems associated with this approach. In deep MLP, early hidden layers don't get trained well. The reason is the error attenuates as it propagates to earlier layers. Gradients are much smaller at deeper levels and training slows significantly. This is called a difussion of gradient. 

The problem is exacerbated by the fact that in fully connected neural network, number of connections rises with $(h-1)x^{4}$, where $x$ is the pixel dimension of the input image (assuming it's a square) and $h$ is the number of network's layers (excluding the output layer and assuming hidden layers and input layer have the same dimensions). For a 30x30 image and network with three hidden layers, we get $3 240 000$ connections. In combination with the diffusion of gradient, the network can't be trained effectively in a finite time. When training the deep MLP, top layers usually learns any task pretty well and thus the error to earlier layers drops quickly as the top layers mostly solve the task. Lower layers never get the opportunity to use their capacity to improve results, they just do a random feature map.

This obstacle led to a crisis in artificial intelligence in the 90's and decline of artificial neural networks and rise of much simpler architectures such as support vector machines (SVM) and linear classifiers. AI research was facing a lot of pessimism. The common practice in computer vision was preprocessing with manually handcrafted feature extractor and learning with simple trainable classifier. Shallow kernel architectures such as SVM was a preferred machine learning architecture. This was very time-consuming and had to be repeated for different tasks.

Luckily during the years, major breakthroughs happened in deep learning. It turned out there are several ways how to build and run deep architectures:

\begin{itemize}
\item Very recent work has shown significant accuracy improvements by "patiently" training deeper multilayer perceptrons with backpropagation using GPGPU (massive general purpose parallel computations on graphic cards).
\item Convolutional neural nets developed by Yann LeCun is the first truly deep architecture. It's the architecture that closely resembles human visual system. It can be trained supervised using the gradient descent method.
\item Geoffrey Hinton developed deep belief networks which use greedy layer-wise training. Each layer is a restricted boltzmann machine and is trained unsupervised. Final outputs are fed to a supervised model.
\item Team around Yann LeCun found how to train unsupervised, sparse convolutional neural nets. This reduced training time and yielded better performance on tasks where not large enough training set is available.
\end{itemize}

These deep architectures greatly outperform SVM and other shallow kernel topologies. They also require little human input and preprocessing in comparison to SVM. Deep learning is attracting a lot of attention and has been implemented in commercial applications.

\newpage ~ \newpage

\chapter{Convolutional neural networks}

\section{Description}

The following description is based on papers \cite{ann_bouchain} by David Bouchain and \cite{cnn_svm}, \cite{cnn_present}, \cite{cnn_new} by Yann LeCun and others.

Concept of convolutional neural networks (ConvNets) was introduced in 1995 by Yann LeCun and Yoshua Bengio \cite{cnn_origin}. Their work was neurobiologically motivated by the findings of locally sensitive and orientation-selective nerve cells in the visual cortex of a cat \cite{ann_bouchain}. Cats have a high performance visual system close to that of primates, making it a very coveted subject for researches to reveal the functional aspects of this complex system \cite{deep_v1}.

\begin{figure}[!htbp]
\begin{center}
\includegraphics[width=14cm]{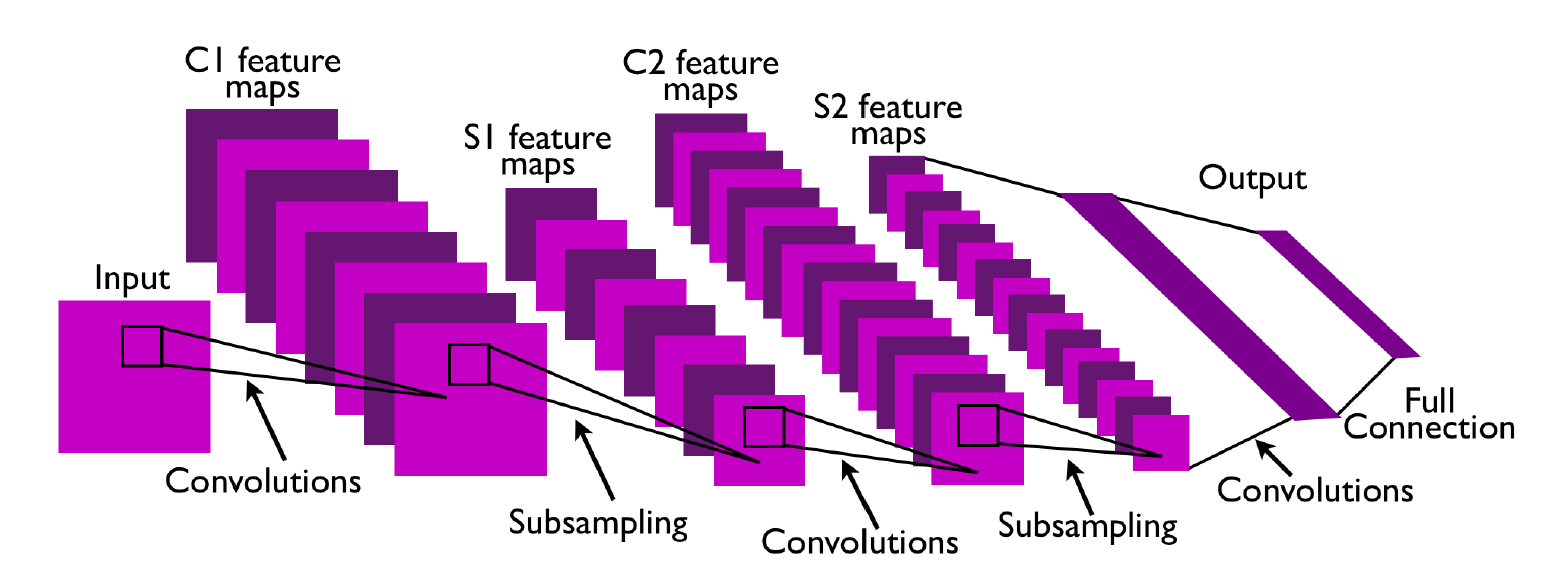}
\caption{Example of a two stage convolutional neural network \cite{cnn_new}.}
\label{pic_cnn}
\end{center}
\end{figure}

Crucial element of ConvNets is restricting the neural weights of one layer to a local receptive field in the previous layer. This receptive field is usually a two dimensional kernel. Each element in the kernel window has a weight independent of that of another element. The same kernel (with the same set of weights) is moved over neurons from the prior layer. Therefore the weights are shared for each feature map, they differ only between different feature maps in each feature map array (layer). Making several feature maps in each layer enables the network to extract multiple features for each location.

\begin{figure}[!htbp]
\begin{center}
\includegraphics[width=14cm]{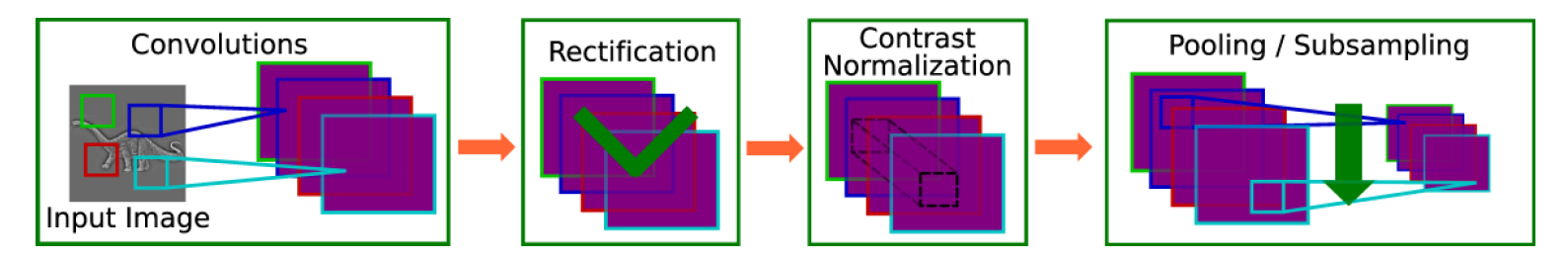}
\caption{Structure of one stage of the ConvNet \cite{cnn_new}.}
\label{pic_stage}
\end{center}
\end{figure}

Convolutional neural network is composed of several feature map arrays clustered into one, two or three stages, followed by a classifier at the end. Each stage serves as a feature extractor. The clasifier is usually a fully connected neural network which computes the product of the feature vector. Each stage is made by filter bank layer, nonlinearity layer and feature pooling layer.

\begin{itemize}
\item \textbf{Filter bank layer} - Input is a set of $n_{1}$ two dimensional feature maps of size $n_{2} \times n_{3}$. Each feature map is denoted $x_{i}$ and each neuron $x_{ijk}$. Output is a different set of feature maps  $y_{j}$. A two dimensional trainable kernel $k_{ij}$ connects feature map $x_{i}$ and $y_{j}$. The filter computes

\begin{equation}
y_{j} = b_{j} + \sum_{i}k_{ij} * x_{i}
\label{cnn_discrete}
\end{equation}

where $b_{j}$ is a trainable bias parameter and $*$ operator performs two dimesional discrete convolution. Number of feature maps at the output is determined by how many different convolutional kernels we use. Each kernel can extract different features from the image (edges, etc.).
\item \textbf{Nonlinearity layer} - This is usually a pointwise $\tanh()$ sigmoid function

\begin{equation}
y_{j} = \tanh(x_{i})
\label{cnn_tanh}
\end{equation}

or more preferred rectified sigmoid

\begin{equation}
y_{j} = \left|g_{i} \cdot \tanh(x_{i})\right| 
\label{cnn_rect}
\end{equation}

where $g_{i}$ is a trainable gain parameter. This is followed by a subtractive and divisive local normalization which enforces local competition between neighboring features in a feature map and between features at the same spatial location. The subtractive normalization is made by

\begin{equation}
v_{ijk} = x_{ijk} - \sum_{ipq} w_{pq} \cdot x_{i,j+p,k+q}
\label{cnn_subtr}
\end{equation}

where $w_{pq}$ is a normalized truncated Gaussian weighting window. The divisive normalization is then

\begin{equation}
y_{ijk} = \frac{v_{ijk}}{\mathrm{max}(\mathrm{mean}(\sigma_{jk}),\sigma_{jk})}
\label{cnn_divis}
\end{equation}

where

\begin{equation}
\sigma_{jk} = \sqrt{\sum_{ipq} w_{pq} \cdot v_{i,j+p,k+q}^{2}}
\label{cnn_divis2}
\end{equation}
\item \textbf{Feature pooling layer} - The purpose of this layer is to reduce the spatial resolution of preceding feature maps. This is very beneficial, because it eliminates some position information about features, therefore builds some level of distortion invariance in the representation. The transformation may be simply computation of an average or maximal value over the receptive field. Most common is the \textit{subs} layer doing simple average over the kernel, or more sophisticated \textit{l2pool}. That's a variant of Lp-Pooling, which is a biologically inspired pooling process modelled on complex cells \cite{ach_house}. Its operation is described by equation 

\begin{equation}
O = (\sum\sum I(i,j)^{P} \times G(i,j))^{1/P}
\label{cnn_pool}
\end{equation}

where $G$ is a Gaussian kernel, $I$ is the input feature map and $O$ is the output feature map. It can be imagined as giving an increased weight to stronger features and suppressing weaker features. Two special cases of Lp-Pooling are notable. $P = 1$ corresponds to a simple Gaussian averaging, whereas $P = \infty$ corresponds to max-pooling (i.e only the strongest signal is activated) \cite{ach_house}. 

Some ConvNets use stride larger than one to further reduce the spatial resolution. Pooling is followed by tanh() nonlinearity in most networks, but in some recent models don't.
\end{itemize}

All layers are trained at the same time using supervised gradient based descent described in the previous chapter. In some recent models, a method for unsupervised training has been proposed. It's called a Predictive Sparse Decomposition (PSD) \cite{cnn_psd}.

\section{Advantages}

\begin{itemize}
\item ConvNet implicitly extracts relevant features from the image. It's naturally best architecture for computer vision. That's because it's working on the same priciple as the human visual system. It hierarchically abstracts unique features of the input as the visual cortex. Convolution and subsampling is a native process in the human visual system, which can be proved by various visual tricks \cite{deep_vis_system}.
\item By being a deep architecture and by reducing spatial resolution of the feature map, ConvNets achieve a great degree of shift and distortion invariance. This is something SVM and other shallow architectures completely lack.
\item Convolutional layers have significantly less connections thanks to shared weights concept. Subsampling layers have usually only one trainable weight and one trainable bias. Overall this massively reduced the number of parameters in comparison to multilayer perceptrons. This makes ConvNets exponentially easier to train than MLPs.
\item Unlike SVM and similar shallow kernel template matchers, ConvNets feature extractors are trained automatically and don't have to be manually handcrafted. This greatly saves human time and more closely resembles true artificial intelligence.
\item Unlike multilayer perceptrons, where performance was sensitive to networks parameters, ConvNets perform universally good with various topologies and aren't so sensitive to pretraining normalization. 
\end{itemize}

\section{Achievements}

ConvNet was the first truly deep architecture. Therefore it has long history of sucessful application in OCR and handwriting recognition. However most excitement came from more recent results.

\begin{itemize}
\item AT\&T developed a bank check reading system using ConvNet in the 90's. It was deployed on check reading ATM machines of large US and European banks. By the end of 90's it was reading over 10\% of all checks in the US.
\item Microsoft deployed ConvNets in a number of handwriting recognition and OCR systems for Arabic \cite{ach_arab} and Chinese \cite{ach_asia} characters.
\item Google used a ConvNet in its Google StreetView service. It was a component of algorithms that search and blur human faces and license plates \cite{ach_street}.
\item Video surveillance system based on ConvNets has been deployed in several airports in the US.
\item ConvNets trained unsupervised with sparsifying non-linearity achieved record results in handwritten digits recognition using MNIST dataset with error rate as low as 0.39\%, outperforming all methods previously used \cite{ach_mnist}.
\item ConvNet is the most suitable architecture for object recognition. It achieved the best score on NORB dataset (96x96 stereo images) with error only 5.8\%. Second best architecture (Pairwise SVM) achieved 11.3\% \cite{cnn_present}.
\item It was sucessfully applied to house numbers digit classification. This was similar task as the MNIST dataset - the task was numbers recognition on 32x32 images. However in this case it was an order of magnitude bigger dataset (600000 labeled digits), contained color information and various natural backgrounds. The best used architecture (multi-stage ConvNet trained supervised) obtained accuracy of 95.1\%, very close to the human performance of 98\% \cite{ach_house}.
\item Maybe the most notable application was the traffic signs and pedestrian recognition. Team around Pierre Sermanet and Yann LeCun trained an unsupervised ConvNet using Predictive Sparse Decomposition. They tried using various transformations of the dataset and the best obtained accuracy was an outstanding 99.17\% (others were 98.97\% and 98.89\%). However the most important fact about this experiment was that for the first time, the AI was able to achieve better performace than human (98.81\% in this case) in a typical human-like problem \cite{ach_traffic}. 
\end{itemize}

\newpage ~ \newpage

\chapter{Practical Application}

\section{Preparing the dataset}

One thing I've learned during my previous work with multilayer perceptrons is that quality of the training set is critical. It's the most influential factor in training of neural networks. Not the topology of the network, type of learning algorithm or anything else. Even the most advanced and fine tuned machine learning algorithms are useless if you feed them with improperly selected data that don't describe the underlying pattern accurately. Second most important factor in case of multilayer perceptron was the preprocessing and data representation.

Convolutional neural networks are not so dependent on preprocessing. They can handle data with little or no preprocessing without problems. However, preprocessed or not, the data still have to represent the underlying problem as accurately as possible. Therefore it's no surprise I spent a great deal of time designing the selection algorithm. I downloaded the BOSS survey catalog from \cite{sdss}. It's a 1.6GB fits file containing a giant list of 1.5 million spectra obtained by BOSS spectrograph. 

\begin{table}[!htbp]
\begin{center}
\begin{tabular}{c|r|r}
\textbf{Class} & \textbf{Total} & \textbf{Unique}	\\
\hline
Total		&	1 507 954	&	1 391 792	\\
Galaxies	&	927 844	&	859 322	\\
Quasars		&	182 009	&	166 300	\\
Stars		&	159 327	&	144 968	\\
Sky			&	144 503	&	138 491	\\
Unknown		&	101 550	&	89 003	\\
\end{tabular}
\end{center}
\caption{Amount of spectra in the BOSS survey \cite{sdss}.}
\label{tab_numspec}
\end{table}

The amount of spectra taken in BOSS survey is substantial. I can't utilize them all. BOSS archive with reduced spectra takes 6.59TB of space. That's a magnitude over my capabilities, not mentioning the computational time required for processing all these spectra. I decided to create smaller, but still large enough dataset for my analysis.

\begin{table}[!htbp]
\begin{center}
\begin{tabular}{c|r}
\textbf{Dataset} & \textbf{Size}	\\
\hline
Training	&	31 775	\\
Validation	&	3 103	\\
Testing		&	60 329	\\
\end{tabular}
\end{center}
\caption{Size (number of spectra) of my selected datasets.}
\label{tab_sets}
\end{table}

Each spectrum is described by three numbers that, when combined together, create a unique identifier of the spectrum. It's a number of the spectrographic plate, modified Julian date (MJD) when spectrum was obtained, and a number of the fiber on the spectrographic plate. Each spectrum has its own record in the BOSS catalog. The record includes many flags and parameters describing instrumental response, quality of the spectrum (\texttt{ZWARNING} flag), and products of the spectroscopic pipeline. The parameters we are interested in are:

\begin{table}[!htbp]
\begin{center}
\begin{tabular}{r|p{11cm}}
\textbf{Class} & \textbf{Total}	\\
\hline
\texttt{PLATE}		&	number of the spectrographic plate	\\
\texttt{MJD}		&	modified Julian date when the spectrum was taken	\\
\texttt{FIBERID}	&	number of the optic fiber plugged into the plate	\\
\texttt{SPECBOSS}	&	indicates whether spectrum is primary (the best observation of that location)	\\
\texttt{ZWARNING}	&	a flag indicating whether the spectrum had any problems in the spectroscopic pipeline	\\
\texttt{CLASS}		&	object type classification (galaxy, quasar or star), product of the spectroscopic pipeline	\\
\texttt{Z}			&	estimated redshift, product of the spectroscopic pipeline	\\
\end{tabular}
\end{center}
\caption{Parameters extracted from the BOSS catalog for each spectrum.}
\label{tab_catalog}
\end{table}

I used the catalog to construct as good training set as possible. It had to equally represent every object type. Therefore the ratio of galaxies, quasars and stars in the training set is roughly 1:1:1. This was an easy task. Ensuring the equal distribution of objects by redshift among individual object classes proved to be more challenging. This is important mainly for quasars that span over wide interval of redshifts. Narrowing this distribution is desirable from two reasons. First, spectra are distorted becauses of doppler effect:

\begin{equation}
z = \frac{\lambda_{obsv}}{\lambda_{emit}} - 1
\label{doppler}
\end{equation}

This means spectral lines or any other characteristic elements for the classification are shifted in high $z$ spectra. Second problem is the selection bias. High $z$ quasars are very distant objects. In fact these objects can be as far as the edge of the observable universe. Because they are so far, it is right to assume the objects we observe are the ones that are most bright. If they would be faint, we won't see them at such a great distance. This was scientifically proved. Brightness of the quasar depends on the inclination of its jet towards our line of sight. The most luminous quasars are called blazars and their jets are aimed directly towards us, amplified by the relativistic beaming effect \cite{beaming}. Spectra of blazars are dominated by nonthermal effects such as synchrotron emission and inverse compton scattering at higher energies. So would the distribution of quasars in the training set be skewed towards high redshift, the classification algorithm would consider a typical quasar spectra to look similar to that of high $z$ blazar. Which would be incorrect assumption and classification algorithm can then perform poorly on low $z$ quasar spectra.

\begin{figure}[!htbp]
\begin{center}
\includegraphics[width=14cm]{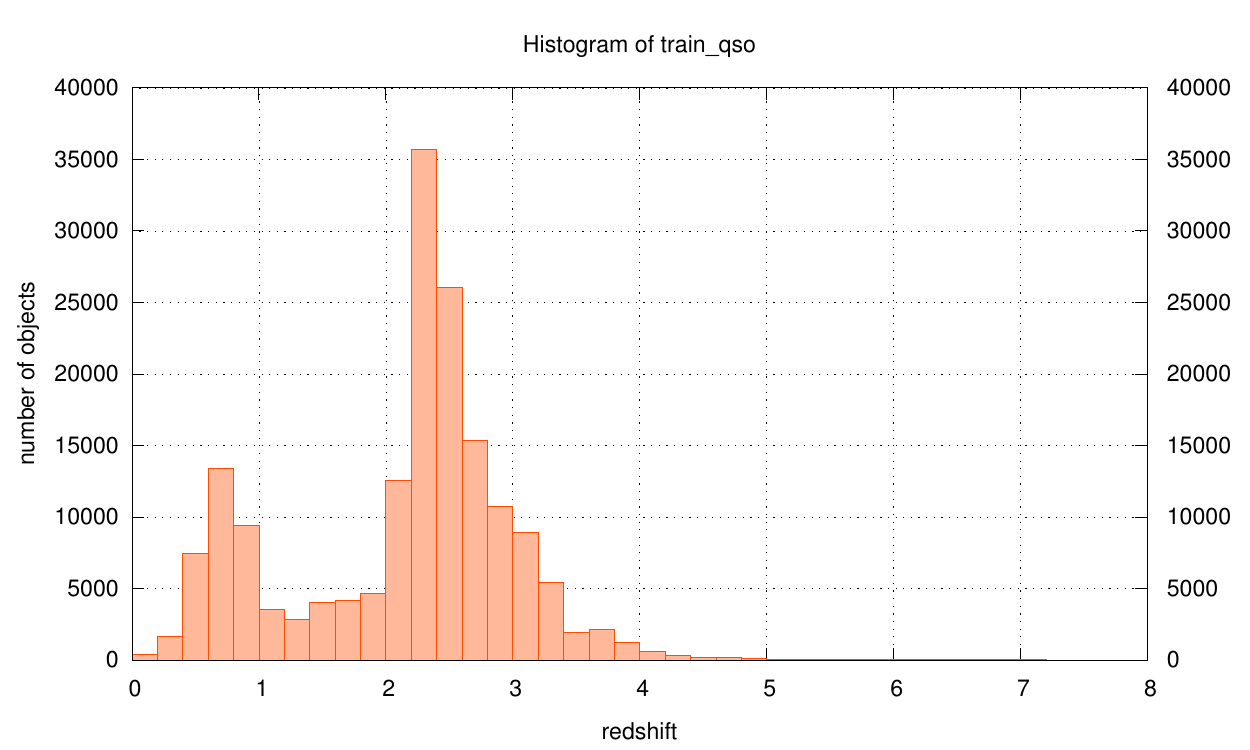}
\caption{Distribution of all the good (\texttt{SPECBOSS}=1 and \texttt{ZWARNING}=0) quasar spectra in the BOSS catalog.}
\label{pic_hist_qso_r}
\end{center}
\end{figure}

\begin{figure}[!htbp]
\begin{center}
\includegraphics[width=14cm]{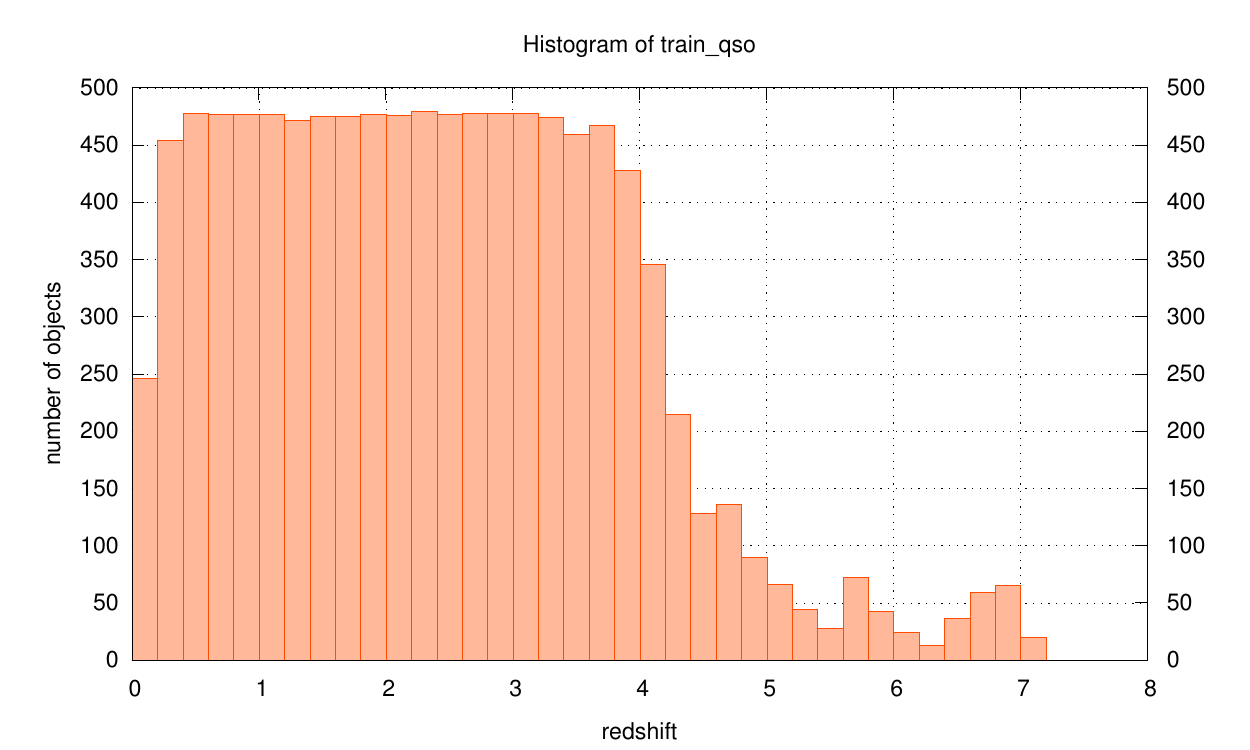}
\caption{Distribution of quasar spectra in the training set.}
\label{pic_hist_qso_s}
\end{center}
\end{figure}

As is shown in figure \ref{pic_hist_qso_r}, the distribution of quasars by redshift in the catalog is highly nonhomogenous. There is tens of thousands of objects in intervals from two to three and zero to one, while only few hundreds of objects in interval from five to seven. I developed a selection algorithm that divides the redshift range into equidistant intervals and required amount of samples for each inteval is computed. The samples are then picked from each interval using modified pseudorandom generator. After this procedure, the distribution of quasars by redshift is in the figure \ref{pic_hist_qso_s}, which are in fact objects that will be included in the training set.

\begin{figure}[!htbp]
\begin{center}
\includegraphics[width=14cm]{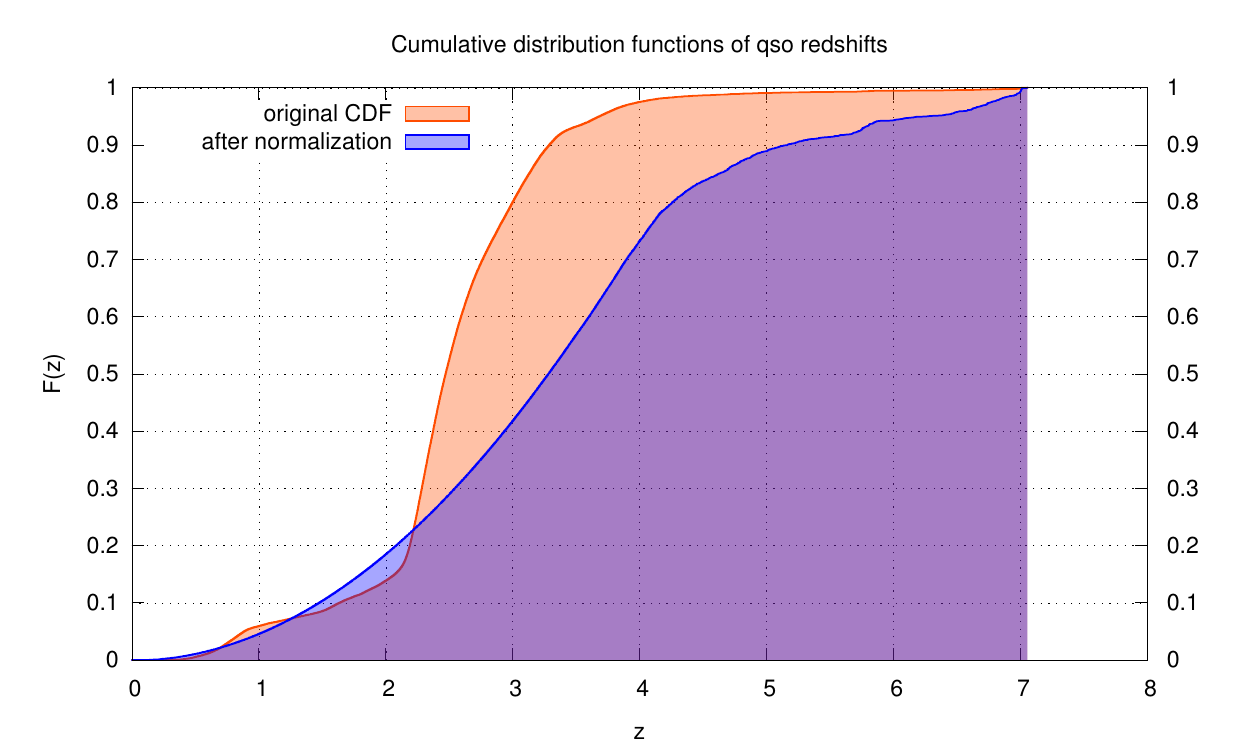}
\caption{Cumulative distribution functions ($z$ is redshift) for all good quasars in the BOSS catalog (red) and for quasars selected into the training set (blue).}
\label{pic_cdf_qso}
\end{center}
\end{figure}

The distribution in figure \ref{pic_hist_qso_s} for objects with redshift over four is not constant because there is not enough objects with so high value of redshift in the BOSS catalog. The difference can be statistically highlighted by comparing cumulative distribution functions for both sets. This was done in figure \ref{pic_cdf_qso}. This proves the selection alorithm is very effective.

\begin{figure}[!htbp]
\begin{center}
\includegraphics[width=12cm]{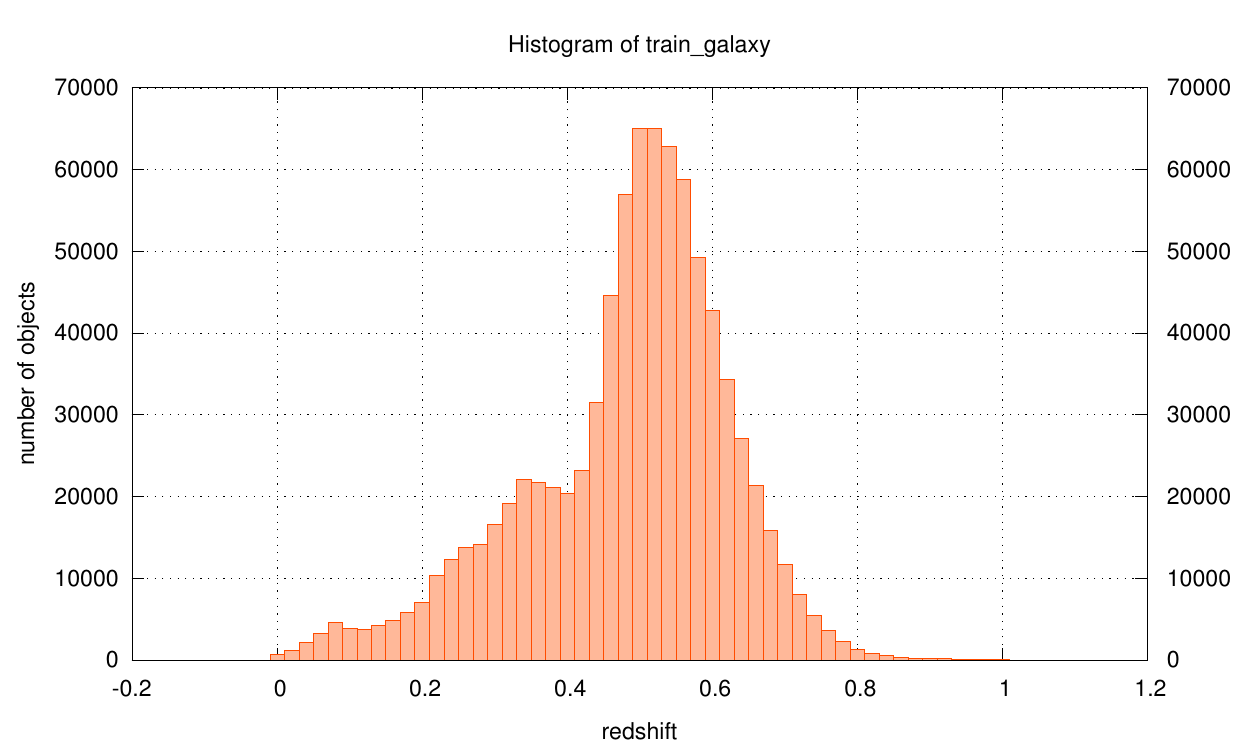}
\caption{Distribution of all the good (\texttt{SPECBOSS}=1 and \texttt{ZWARNING}=0) galactic (non-quasistellar) spectra in the BOSS catalog.}
\label{pic_hist_galaxy_r}
\end{center}
\end{figure}

\begin{figure}[!htbp]
\begin{center}
\includegraphics[width=13cm]{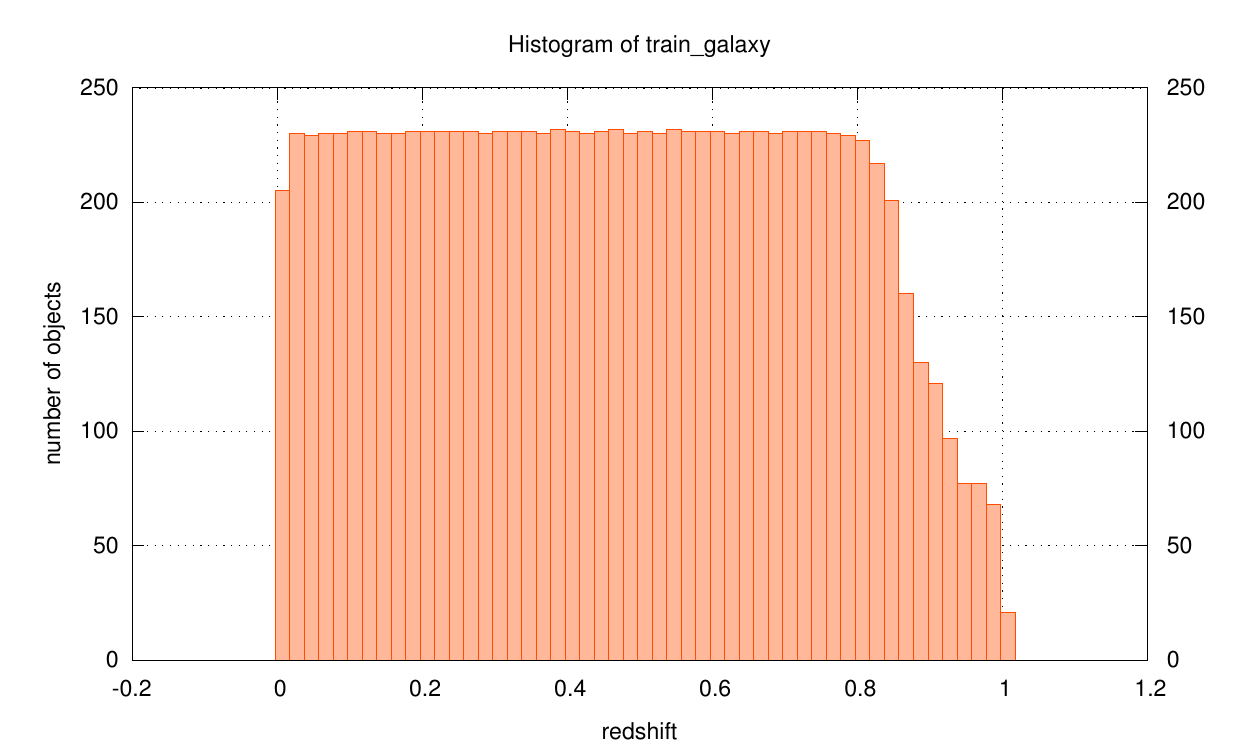}
\caption{Distribution of galactic (non-quasistellar) spectra in the training set.}
\label{pic_hist_galaxy_s}
\end{center}
\end{figure}

\begin{figure}[!htbp]
\begin{center}
\includegraphics[width=13cm]{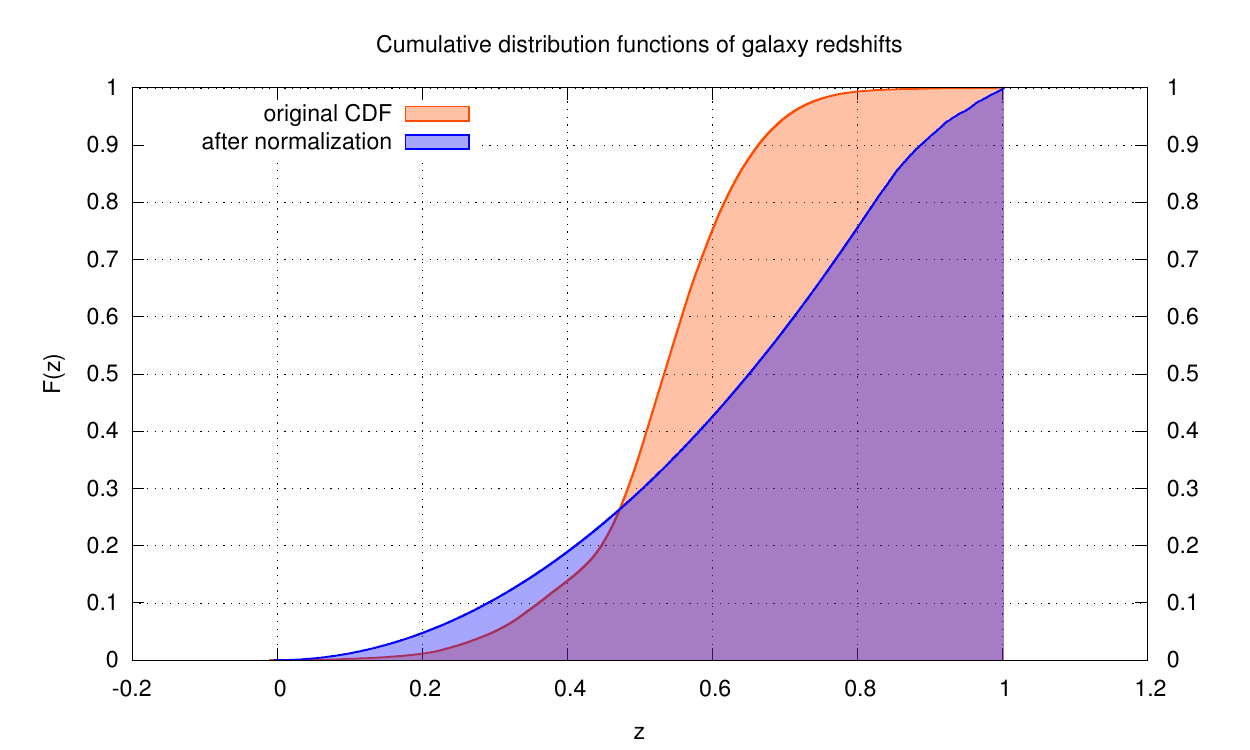}
\caption{Cumulative distribution functions ($z$ is redshift) for all good galaxies (non-qso) in the BOSS catalog (red) and for galaxies (non-qso) selected into the training set (blue).}
\label{pic_cdf_galaxy}
\end{center}
\end{figure}

\begin{figure}[!htbp]
\begin{center}
\includegraphics[width=13cm]{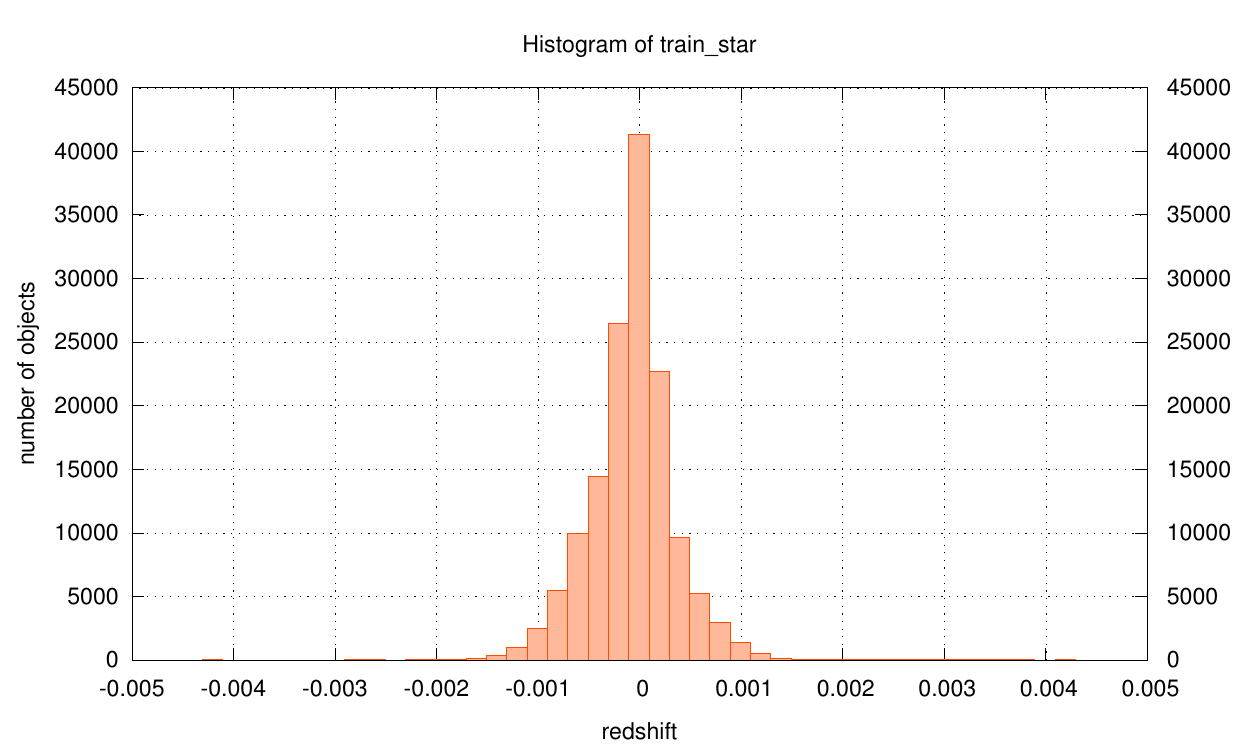}
\caption{Distribution of all the good (\texttt{SPECBOSS}=1 and \texttt{ZWARNING}=0) stellar spectra in the BOSS catalog.}
\label{pic_hist_star_r}
\end{center}
\end{figure}

\begin{figure}[!htbp]
\begin{center}
\includegraphics[width=13cm]{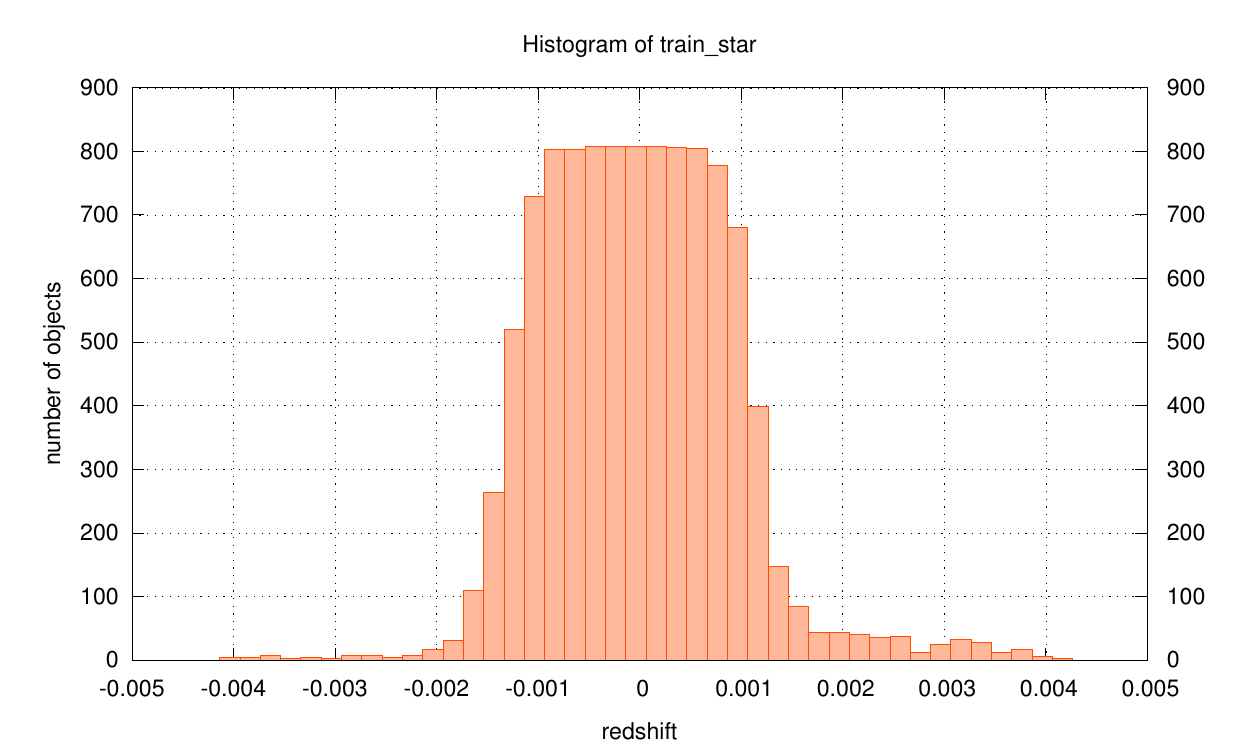}
\caption{Distribution of stellar spectra in the training set.}
\label{pic_hist_star_s}
\end{center}
\end{figure}

\begin{figure}[!htbp]
\begin{center}
\includegraphics[width=13cm]{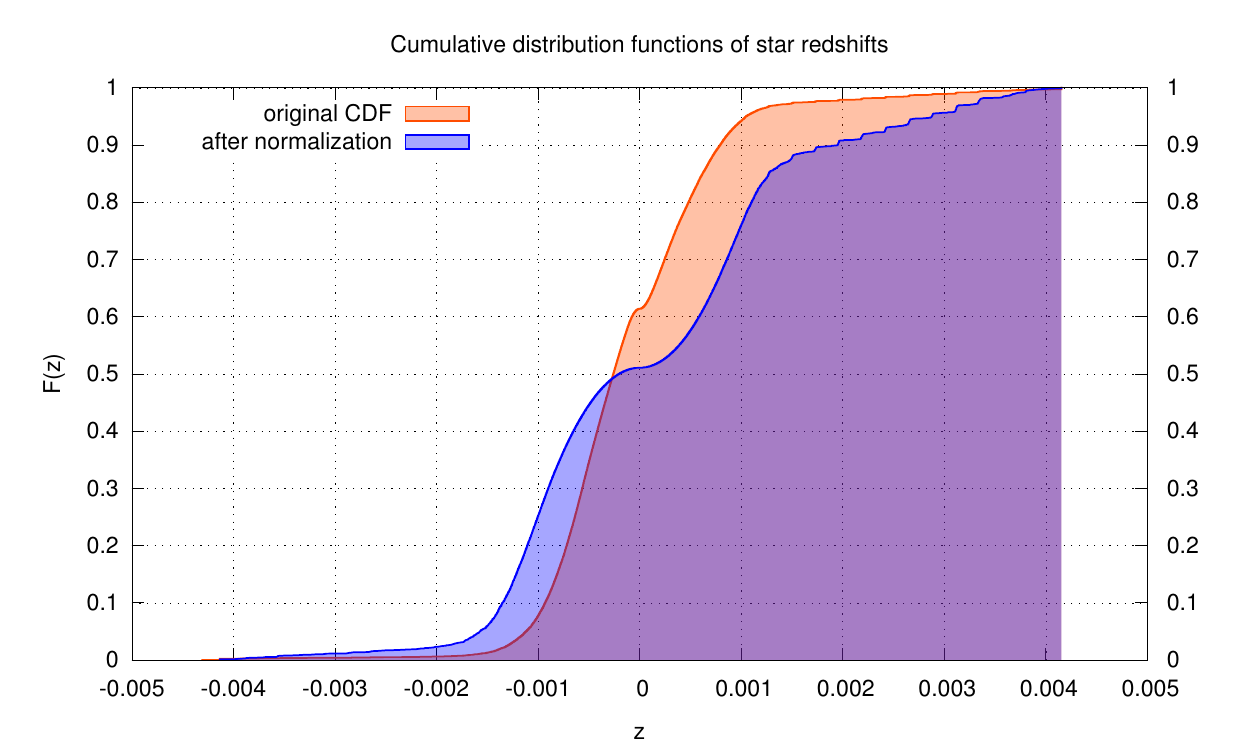}
\caption{Cumulative distribution functions ($z$ is redshift) for all good stars in the BOSS catalog (red) and for stars selected into the training set (blue).}
\label{pic_cdf_star}
\end{center}
\end{figure}

Unlike quasars, galactic (non-quasistellar) and stellar objects are distributed over much narrower interval of redshifts. Applying the selection algorithm on them doesn't bring such an advantage as in quasar's case, but even though it's still beneficial. Corresponding charts for non-quasistellar galaxies are in figures \ref{pic_hist_galaxy_r}, \ref{pic_hist_galaxy_s}, \ref{pic_cdf_galaxy} and for stars in figures \ref{pic_hist_star_r}, \ref{pic_hist_star_s}, \ref{pic_cdf_star}.

The selection algorithm is applied also for creation of validation and testing sets, although quality of these sets is not so important.

\clearpage
\section{Preprocessing}

After creation of training, validation and testing datasets, the spectra corresponding to records in datasets were downloaded from the Science Archive Server \cite{sas}. Each downloaded spectrum is stored as a table in a separate fits file. This means I got 95207 fits files to deal with. Although there are ways how to deal directly with the fits files (using routines from the \texttt{CFITSIO} library), I decided to extract required data into simple text files, so I can work then with the spectra using standard \texttt{FILE} structure from \texttt{stdio} header file. Extraction of the data from fits files was performed using heatools package from HEAsoft software pack \cite{heasoft}. I have positive experience with heatools from my bachelor thesis, where I successfully used it for handling huge multi-hdu fits files containing high energy astrophysical data \cite{bakal}.

\begin{figure}[!htbp]
\begin{center}
\includegraphics[width=14cm]{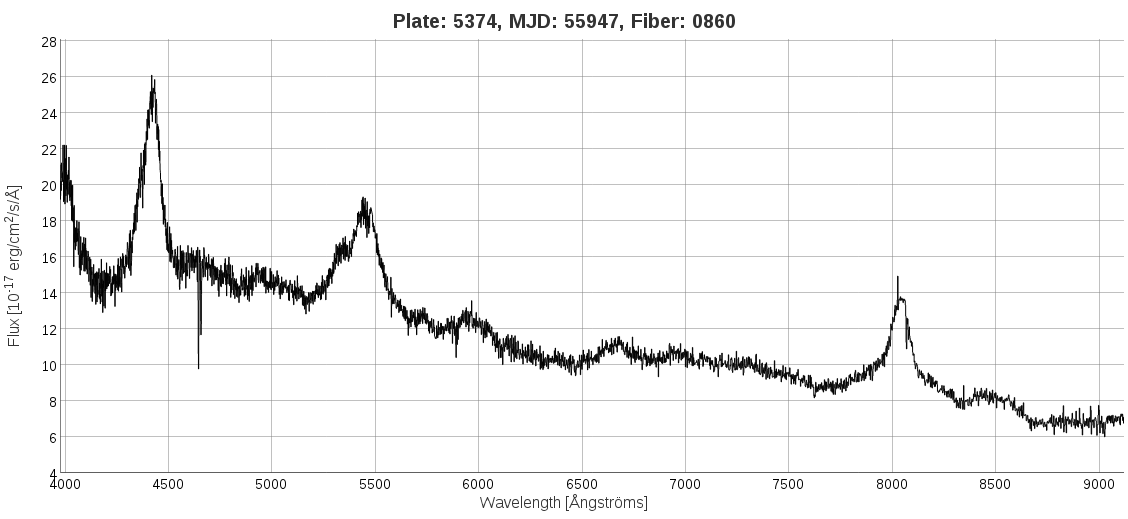}
\caption{Reduced spectrum of quasar 5374-55947-0860 (PLATE-MJD-FIBERID) as a one dimensional vector of intensities.}
\label{pic_spec1d}
\end{center}
\end{figure}

\begin{figure}[!htbp]
\begin{center}
\includegraphics[width=2cm]{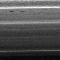}
\caption{The same spectrum of quasar 5374-55947-0860 normalized and converted into a two dimensional 60x60px image.}
\label{pic_spec2d}
\end{center}
\end{figure}

I've written two preprocessing routines. The first is responsible for extraction of spectra, reduction, binning and filtering. The spectra are stored as a one dimensional vector of flux at corresponding wavelenghts. Flux is calibrated in $10^{-17} erg/s/cm^{2}/Ang$ and wavelength in $log_{10}\ Ang$. I decided to arbitrarily reduce the spectra into interval of [3.60000; 3.96000] which corresponds to wavelengths from 3981 \AA\ to 9120 \AA. This is beneficial because edges of most spectra are usually very noisy and doesn't posess any meaningful information for the classification. I incorporated a binning algorithm, so the spectra can fit into a square matrix. Default specral vector has 3601 elements after the reduction and most common binning I used was a 60x60 matrix (effectively no binning). Filtering subroutine searches for spectra that are somehow impaired and removes them from the dataset. 

The spectra are one dimensional vectors, while input into the ConvNet has to be a two dimensional matrix (image). Second routine transforms the spectra from one dimensional vector into two dimensional matrix, normalizes elements of this matrix and finally transforms the matrix into an image. I had several dimensionality transformation algorithms in mind, but decided to simply add elements line after line into the matrix. I was inspired by the way CRT monitors rasterize image on the shadow mask. The output matrices of intensities were normalized using linear normalization and 8-bit grayscale encoding. Finally the product was then converted into an actual PNG image using \texttt{imagemagick} scripting tools, which is the fastest and most reliable approach I have found. The process is illustrated in figures \ref{pic_spec1d} and \ref{pic_spec2d}.

\section{Designing the ConvNet}

I've used EBLearn library written by Pierre Sermanet and Yann LeCun \cite{eblearn} in my scripts and programs for training and testing convolutional neural networks. I deployed two convolutional neural network architectures, both of them designed by Yann LeCun.

\begin{figure}[!htbp]
\begin{center}
\includegraphics[width=14cm]{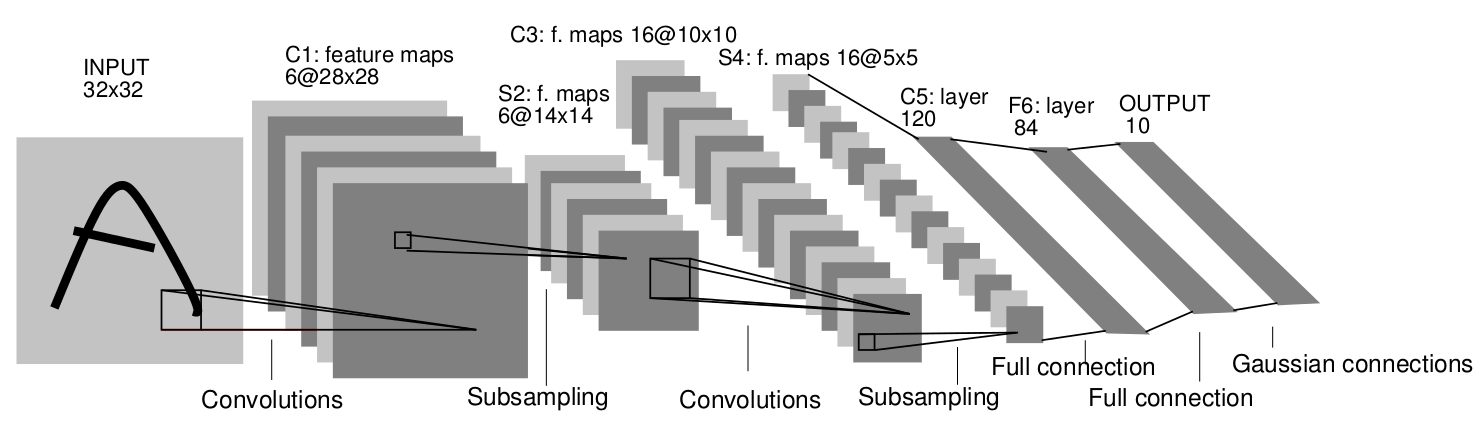}
\caption{LeNet-5 topology used in handwriting recognition on MNIST dataset \cite{lenet5}.}
\label{pic_lenet5}
\end{center}
\end{figure}

The first is LeNet-5 displayed in figure \ref{pic_lenet5}. It's the architecture that was deployed on MNIST dataset, which are 32x32 images of handwritten numbers. I intend to use 60x60 images as an input. The network has to be redesigned in order to incorporate bigger input. I have a lot of experience with designing multilayer perceptron feedforward neural nets. MLP is a two dimensional topology, where it's quite easy to make sure layers and connections match together. ConvNet, on the other side, is a four dimensional body with some layers not fully connected together, where it's obviously much more complicated to design the topology. It took me some time to fully understand how to compute it. In case the input image is smaller than the network is designed for, padding the image to required size is the best practice. When the image is larger, the only way is adjusting correctly size of kernels, or choosing completely different architecture. In case of 60x60 input image and LeNet-5 architecture, I found out the optimal way is enlarging the last convolution kernel between layers S4 and C5 from 5x5 to 13x13 and keeping kernels in previous layers the same. Enlarging all kernels a little (by which way the architecture can also work) doesn't yield as good performance as massively adjusting only the last convolutional kernel.

\begin{figure}[!htbp]
\begin{center}
\includegraphics[width=14cm]{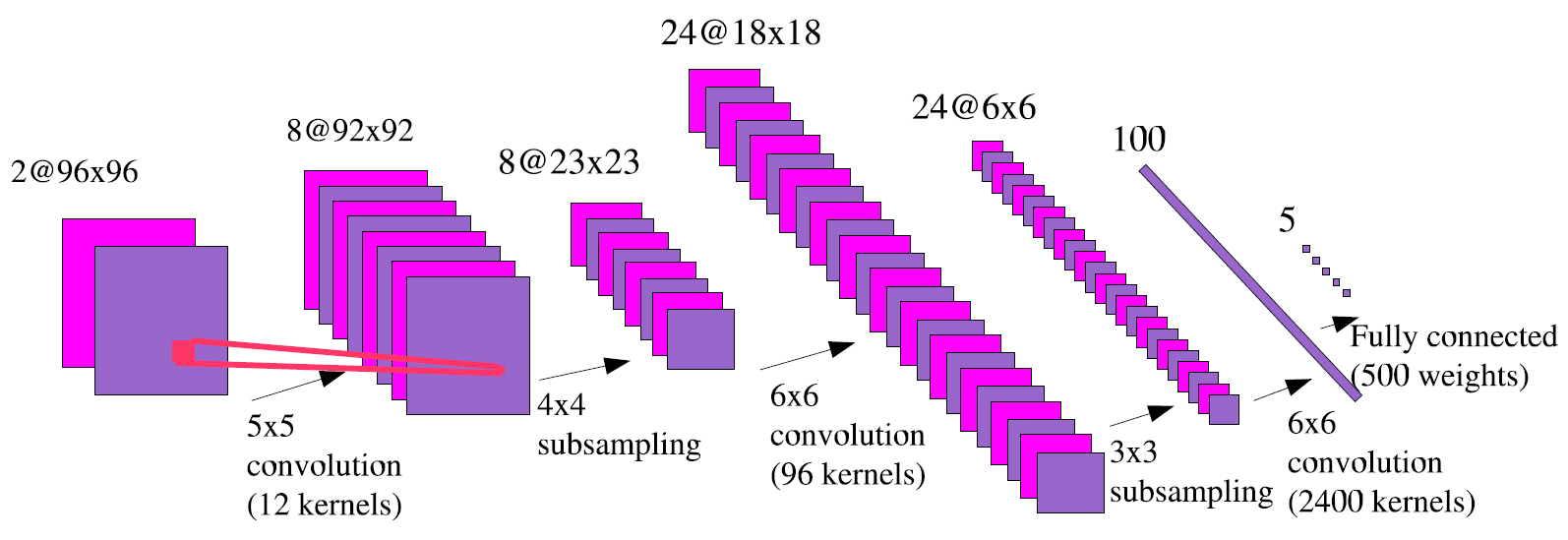}
\caption{LeNet-7 topology used in stereo image recognition on NORB dataset \cite{cnn_svm}.}
\label{pic_lenet7}
\end{center}
\end{figure}

Another way how to classify 60x60 images is using completely different architecture, such as LeNet-7. The problem is that documentation for EBLearn library is quite poor and there is no config file for this architecture anywhere. Therefore I was forced to try to build such an architecture myself. Luckily I was successful. Only difference between my architecture and the LeNet-7 displayed in figure \ref{pic_lenet7} is that mine is using only one image as an input, convolution kernel between layers S2 and C3 is 5x5 instead of 6x6, and subsampling kernel between layers C3 and S4 is 2x2 instead of 3x3. Obviously the dimensions of all feature maps are different than in figure \ref{pic_lenet7}. But besides that the topology is similar, with the same number of feature maps in each layer, the same connection matrices and therefore same abstraction capabilities. The expectation was, that this more massive topology would perform better on my 60x60 images, than more lightweight LeNet-5.

\newpage ~ \newpage

\chapter{Results}
\section{Training}

Before the training could happen, I had to transform training and validation datasets into MAT files required by EBLearn \texttt{train} routine. This was done using \texttt{dscompile} routine. Crucial part of training is a creation of configuration file. It specifies how the architecture has to be built, all required parameters and paths to EBLearn routines and datasets. It took me a lot of time until I fully undestood how to make it, because the documentation provided with EBLearn library is insufficient.

First choice for training was modified LeNet-5 architecture using average subsampling (\textit{subs}) in feature pooling layers, subtractive normalization (\textit{wstd}) in normalization layers and tanh function in nonlinearity layers. ConvNet was trained for 20 epochs (iterations) on training dataset composed of 31775 spectra (galaxies, quasars and stars in ratio roughly 1:1:1) and crosschecked on validation dataset composed of 3103 spectra (galaxies, quasars and stars in ratio roughly 1:1:1), as specified in table \ref{tab_sets}. 

\begin{figure}[!htbp]
\begin{center}
\includegraphics[width=13.9cm]{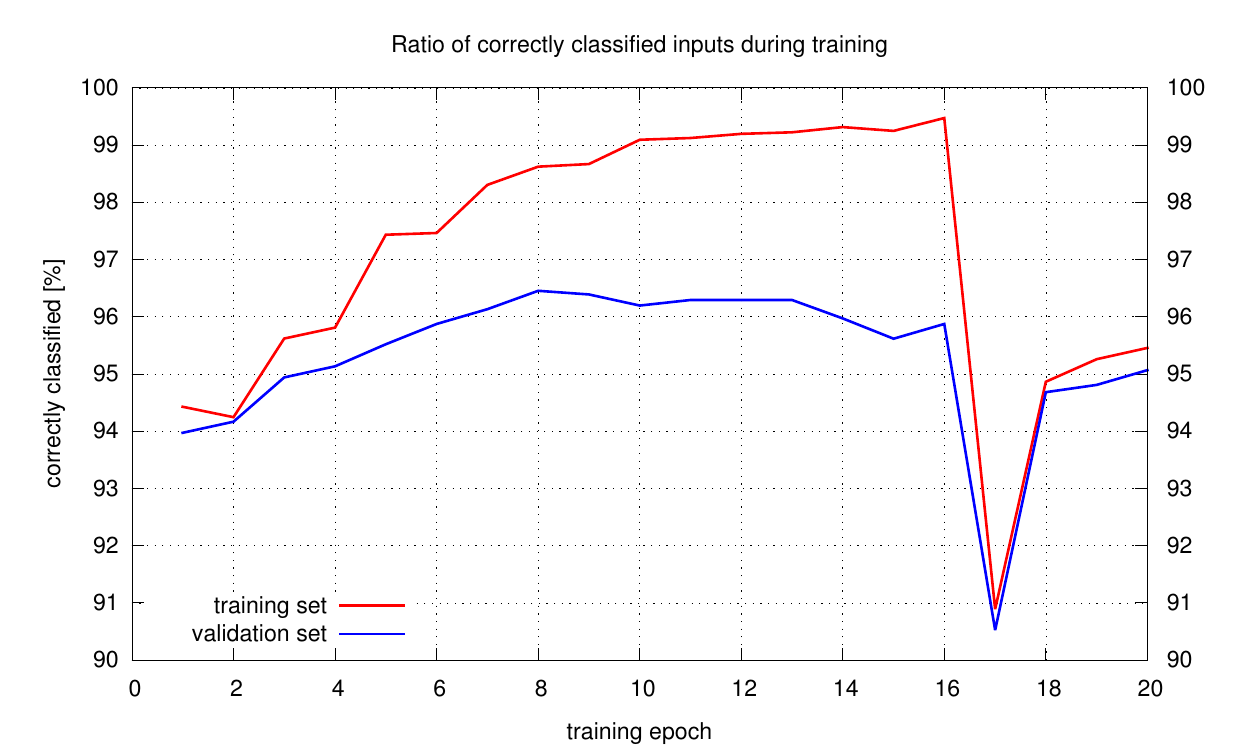}
\caption{Training LeNet-5 architecture (\textit{subs} pooling) for 20 epochs.}
\label{pic_le5_subs_i20}
\end{center}
\end{figure}

\begin{figure}[!htbp]
\begin{center}
\includegraphics[width=13.9cm]{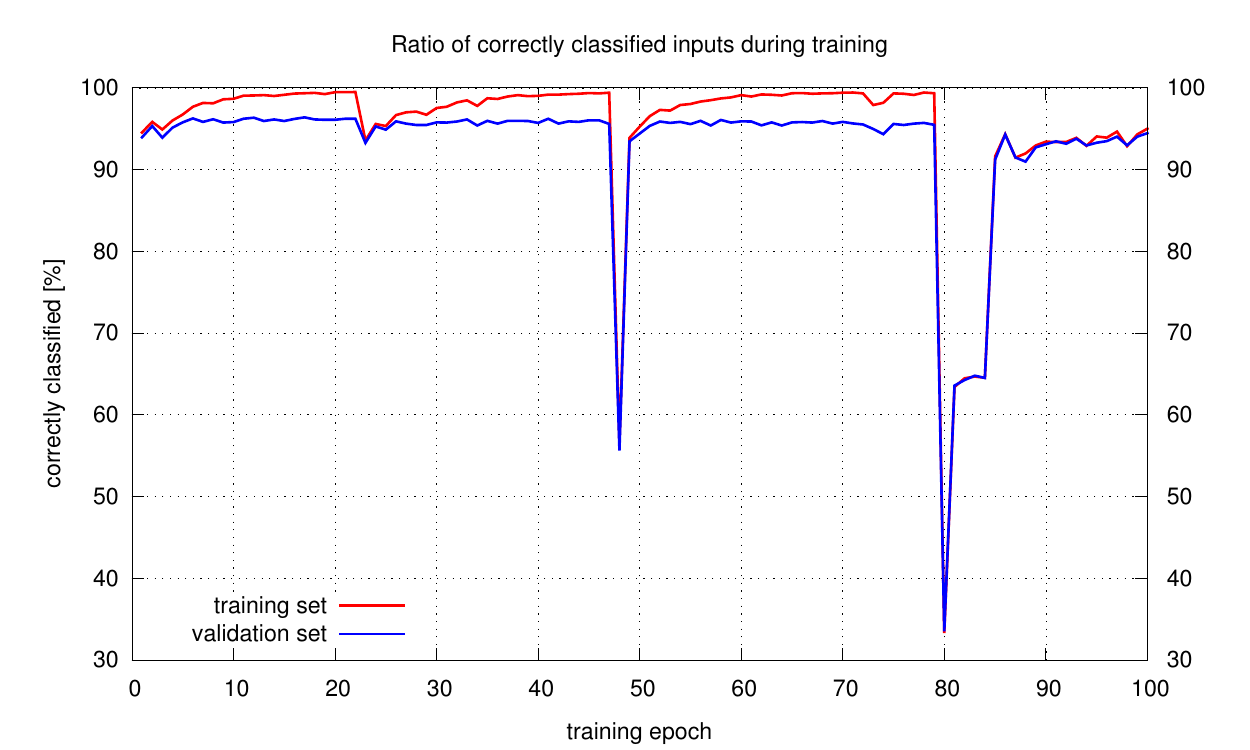}
\caption{Training LeNet-5 architecture (\textit{subs} pooling) for 100 epochs.}
\label{pic_le5_subs_i100}
\end{center}
\end{figure}

The result is in figure \ref{pic_le5_subs_i20}. Considering it's the first shot, results are very good. Success rate on validation set gets over 96\% during training. The problem is that in the 17th epoch the rate collapses. This is not a coincidence. When the training is extended for 100 epochs it became obvious this collapse is a periodical event repeating roughly every 25 epochs. This is displayed in figure \ref{pic_le5_subs_i100}.

\begin{figure}[!htbp]
\begin{center}
\includegraphics[width=13.9cm]{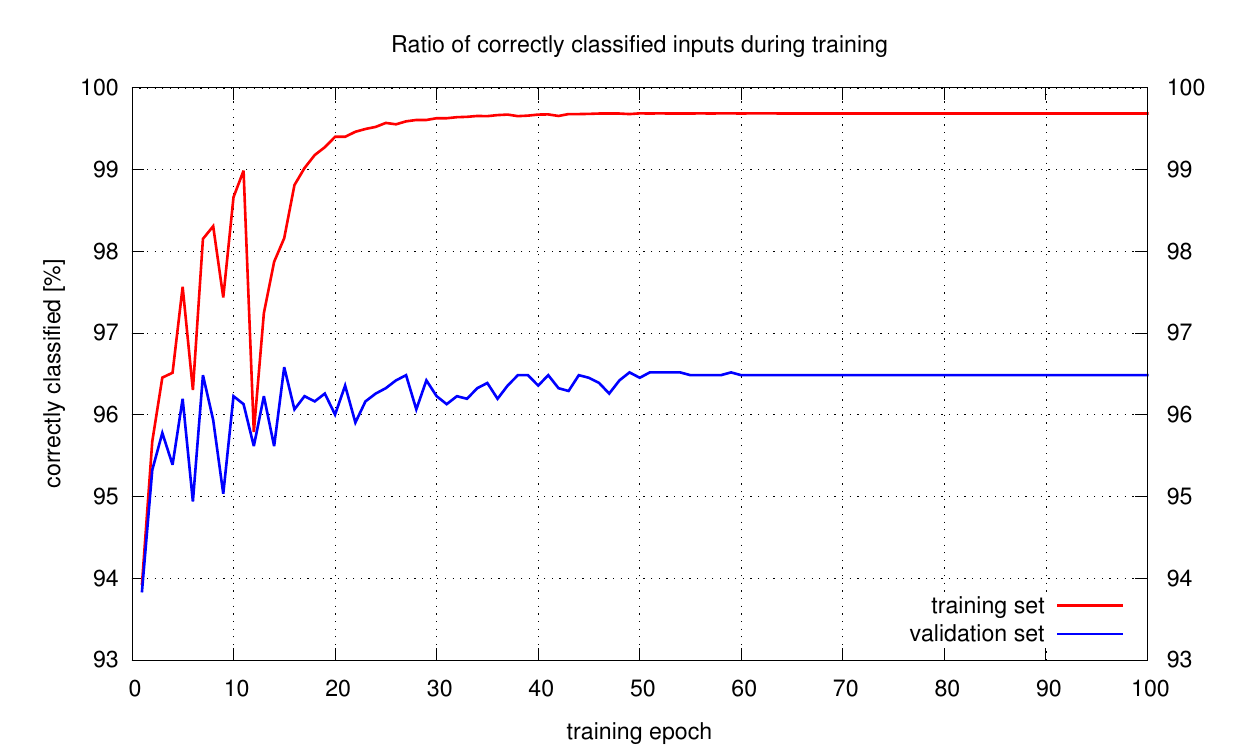}
\caption{Training LeNet-5 architecture (\textit{subs} pooling, eta decay) for 100 epochs.}
\label{pic_le5_subs_dec_i100}
\end{center}
\end{figure}

Based on my experience with multilayer perceptrons, I suspected this instability to be somehow related to the learning rate (eta). Therefore I incorporated a learning rate decay into the configuration file. My theory proved to be correct as can be seen in the figure \ref{pic_le5_subs_dec_i100}. Training was stable and success rate for validation dataset converged at 96.49\%. 

\begin{figure}[!htbp]
\begin{center}
\includegraphics[width=13.9cm]{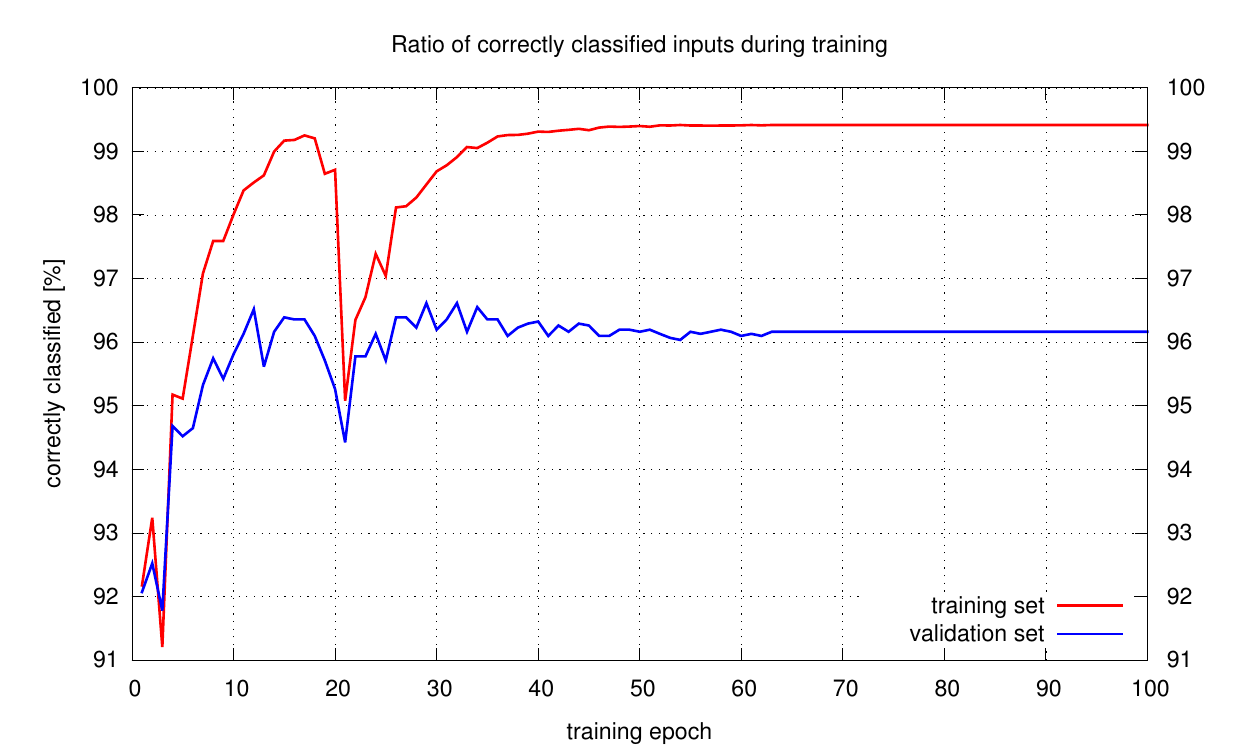}
\caption{Training LeNet-5 architecture (\textit{l2pool} pooling, eta decay) for 100 epochs.}
\label{pic_le5_l2pool_dec_i100}
\end{center}
\end{figure}

I also tried \textit{l2pool} alogithm instead of \textit{subs} in feature pooling layers. The convergence seems to be slower (figure \ref{pic_le5_l2pool_dec_i100}), but success rate at the end (96.16\%) was close to that with \textit{subs} layers.

\begin{figure}[!htbp]
\begin{center}
\includegraphics[width=13.9cm]{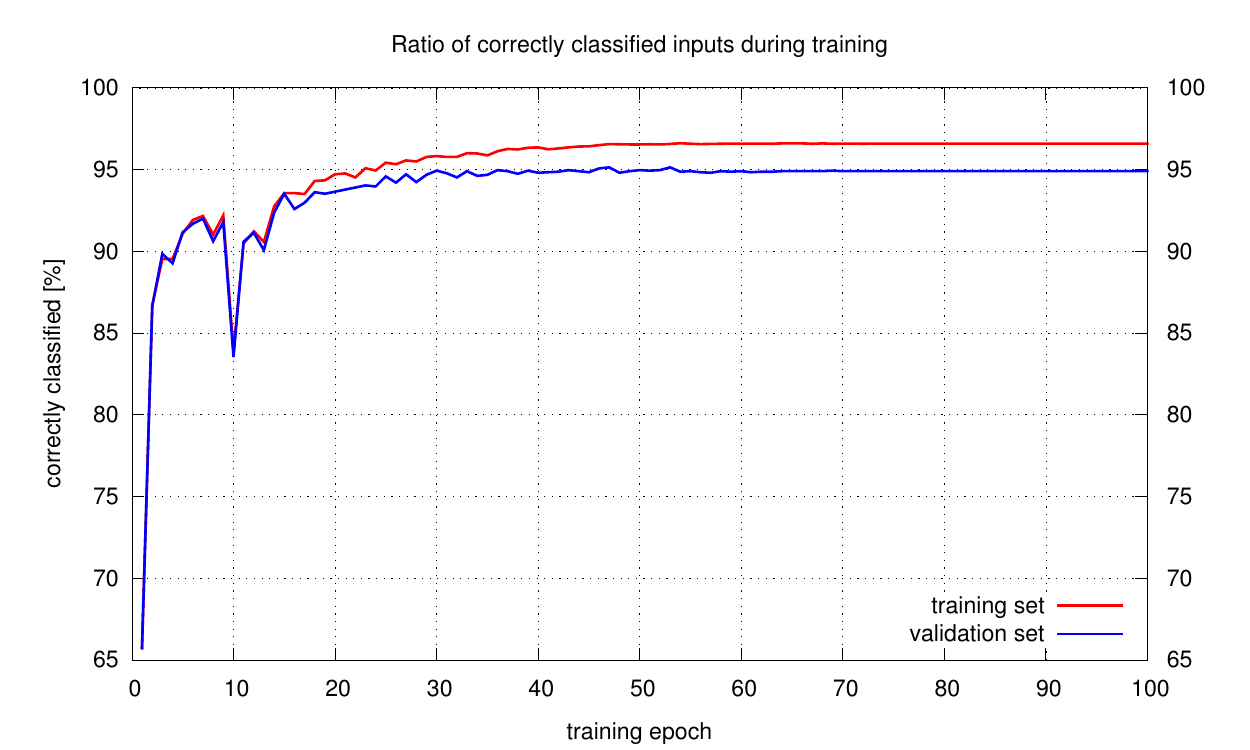}
\caption{Training LeNet-7 architecture (\textit{subs} pooling, eta decay) for 100 epochs.}
\label{pic_le7_subs_dec_i100}
\end{center}
\end{figure}

I was eager to find out how my modified LeNet-7 architecture would perform. I was expecting it to be better than LeNet-5, because it has more feature maps in each layer and was originally designed for bigger images (96x96) than LeNet-5 (32x32). Surprisingly, this assumption proved to be false as can be seen in figure \ref{pic_le7_subs_dec_i100}. The convergence is slower and success rate at the end of training is "only" 94.91\%. Unfortunatelly I couldn't figure out how to train LeNet-7 architecture with \textit{l2pool} pooling due to unspecified error, so only the \textit{subs} variant is included.

\begin{figure}[t]
\begin{center}
\includegraphics[width=13.9cm]{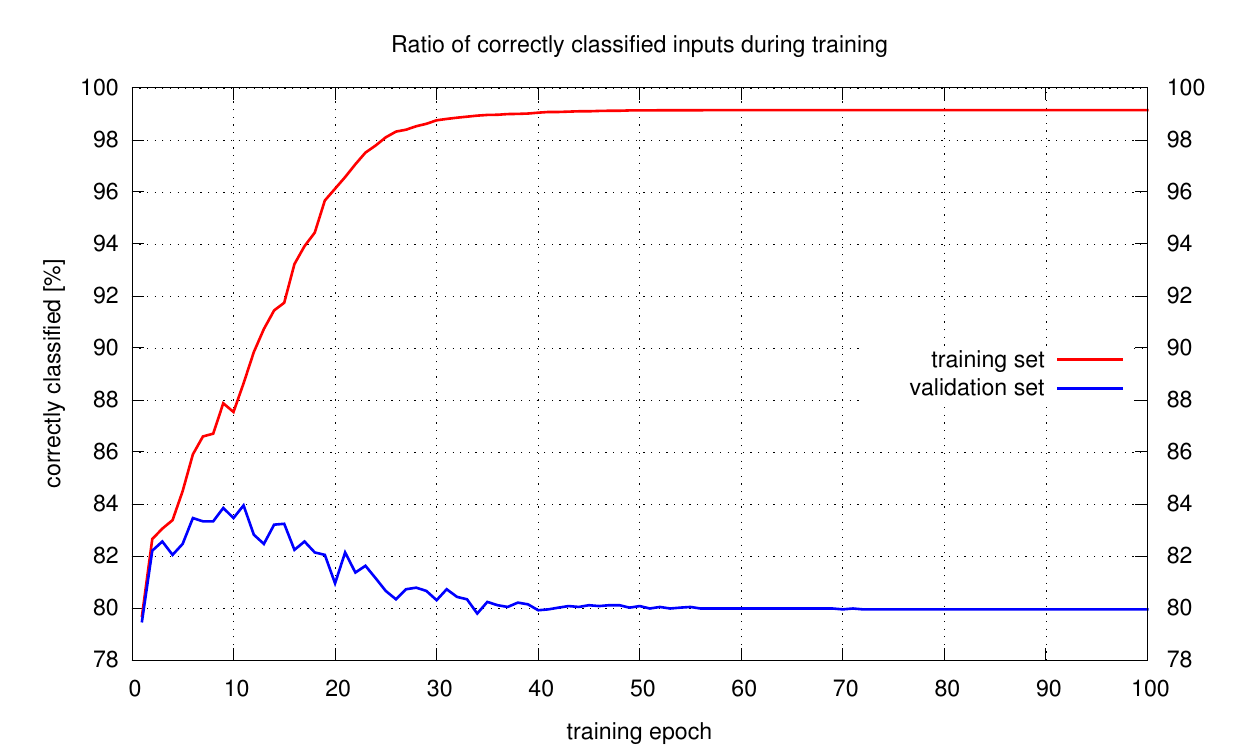}
\caption{Training LeNet-5 architecture (\textit{subs} pooling, eta decay) for 100 epochs using binned spectra.}
\label{pic_le5_subs_dec_i100_28x28}
\end{center}
\end{figure}

One thing I want to clarify is why it's better to use unbinned spectra. Preprocessing program I developed transformed one dimensional reduced spectral vector with 3601 elements into two dimensional 60x60 matrix (which was then transformed into an image). To prove my thesis I created alternative datasets, composed of the same spectra, but binned into 28x28 matrix (image). I modified the configuration file and trained the best performing LeNet-5 architecture with \textit{subs} pooling and eta decay using these binned spectra datasets. Only thing I had to change in the architecture was size of the last subsampling kernel, which had to be shrinked to 5x5 (from 13x13 used for 60x60 images). Training process is visualized in figure \ref{pic_le5_subs_dec_i100_28x28}. Performance is significantly worse than with the unbinned spectra. 

This result has one important implication. Binning the spectra effectively blurred spectral lines, meaning the ConvNet was making classification based on shape of the continuum, rather than position/shape of spectral lines. Significantly worse performace on binned spectra implies the core features ConvNet is exploiting for classification are the spectral lines and not the continuum.

Overall performance for both architectures and different training setups is summed up in table \ref{tab_train}. Two success rates for each variant are included, first is the best achieved during training, and second is the rate at the end of training.

\begin{table}[!htbp]
\begin{center}
\begin{tabular}{c|c|c|c|c|c}
\multirow{2}{*}{\textbf{Input}}	&	\multirow{2}{*}{\textbf{Architecture}}	&	\multirow{2}{*}{\textbf{Pooling}}	&	\multirow{2}{*}{\textbf{Eta decay}}	&	\multicolumn{2}{c}{\textbf{Success rate}}	\\
\cline{5-6}
	&	&	&	&	\textbf{Best achieved}	&	\textbf{At the end}	\\
\hline
60x60	&	LeNet-5		&	subs	&	no	&	96.39\%	&	94.46\%	\\
60x60	&	LeNet-5		&	subs	&	yes	&	96.58\%	&	96.49\%	\\
60x60	&	LeNet-5		&	l2pool	&	yes	&	96.62\%	&	96.16\%	\\
60x60	&	LeNet-7		&	subs	&	yes	&	95.13\%	&	94.91\%	\\
28x28	&	LeNet-5		&	subs	&	yes	&	83.95\%	&	79.95\%	\\
\end{tabular}
\end{center}
\caption{Summarized performace during training. Expressed success rates are always for validation sets.}
\label{tab_train}
\end{table}

\section{Classification}
Next phase was an application of trained networks on the large testing dataset, composed of 60329 spectra. This proved to be one of the greatest obstacles I had to overcome. The problem was that EBLearn has a \texttt{detect} routine for testing. However this routine isn't suitable for classification of whole images. It's best suited for detection of particular patterns in images. For instance when somebody wants to detect human faces in a photo. Some people in EBLearn support forum recommend to adjust bounding boxes in configuration file. I tried this approach, but even then the \texttt{detect} routine results weren't reliable. Success rate was small and results were very unstable. Another fact I found out in the support forum was that there was a \texttt{classify} routine in previous versions of the library, but it was later discontinued and removed.

Only way how to overcome this, was to write my own classification routine from scratch. I found out how to modify the configuration file in such a way, that it forced the \texttt{train} routine to disable the training subroutine and perform only testing on validation set with a saved network configuration. I replaced the validation dataset with the testing dataset and masked it to pretend to be the validation dataset. This way I could successfully perform classification on the testing dataset using already trained network configuration. 

But this approach had one considerable weakness. The \texttt{train} routine isn't designated for testing. When a testing dataset is fed into the routine, the output is only the total sucess rate and sucess rates for each class (galaxy, qso, star). We won't obtain actual classifications, i.e. into which class each object from the dataset belongs. To overcome this I invented a trick. I didn't fed the testing dataset at once into the routine, but rather wrote a feeding script that takes one object from the dataset after another. The script run the object through the \texttt{dscompile} routine and then feed it into the \texttt{train} routine, which is forced to work in classification mode using the method described above. As I said before, I get no additional information from the \texttt{train} routine, but success rate for each class. Because input is only one object, the success rate for one class has to be 100\% and 0\% for two remaining classes. The object has to belong into one of those three classes (galaxy, qso, star). Obviously the class that's assigned the 100\% success rate is the class into which the object conclusively belongs to. The script also has to remember what's the object's identification (PLATE, MJD, FIBERID) during each run, because this information is otherwise lost after the data goes through the \texttt{dscompile} routine. Object's assigned class is then compared with the true class specified in the BOSS catalog. 

My feeding script together with various support scripts is combined into my own \texttt{classify.sh} routine that fully substitutes the original EBLearn \texttt{classify} routine which is no longer available. I ran the testing dataset through this routine for each architecture trained in the previous section. I obtained two lists for each run. The first with identifiers of correctly classified objects and the second with incorrectly classified objects, together with classes assigned by the ConvNet and those from the BOSS catalog. By comparing these two lists I expressed ConvNet's performance on the testing dataset for each architecture.

\begin{table}[t]
\begin{center}
\begin{tabular}{c|c|c|c|c|c}
\multirow{2}{*}{\textbf{Input}}	&	\multirow{2}{*}{\textbf{Architecture}}	&	\multirow{2}{*}{\textbf{Pooling}}	&	\multirow{2}{*}{\textbf{Eta decay}}	&	\multicolumn{2}{c}{\textbf{Success rate}}	\\
\cline{5-6}
	&	&	&	&	\textbf{Best achieved}	&	\textbf{At the end}	\\
\hline
60x60	&	LeNet-5		&	subs	&	no	&	94.78\%	&	93.54\%	\\
60x60	&	LeNet-5		&	subs	&	yes	&	94.72\%	&	94.71\%	\\
60x60	&	LeNet-5		&	l2pool	&	yes	&	94.81\%	&	94.89\%	\\
60x60	&	LeNet-7		&	subs	&	yes	&	93.95\%	&	93.94\%	\\
28x28	&	LeNet-5		&	subs	&	yes	&	85.16\%	&	81.77\%	\\
\end{tabular}
\end{center}
\caption{Summarized performace of trained ConvNet architectures on the testing dataset (60329 objects). Performace is expressed for two classification runs. The first using ConvNet's best achieved configuration (measured by success rate on the validation set) and the second using network's configuration at the end of training.}
\label{tab_test}
\end{table}

Results show the LeNet-5 architecture using unbinned spectra is undoubtedly the best configuration. The learning rate decay carries indisputable advantage. Although it doesn't have significant impact on the best achieved success rate, it removes collapses during training, speeds up training convergence and most importantly the quality of trained ConvNet doesn't depend so heavily on the training epoch. The variant with \textit{l2pool} feature pooling algorithm outperformed the \textit{subs} pooling on both validation and testing datasets. And taking \textit{l2pool} more sophisticated nature into account, it shall be considered superior against \textit{subs} pooling. However it would be better to examine this claim more deeply, because the absolute difference in performance between the \textit{l2pool} and \textit{subs} configurations may be within statistical errors.

\section{Examples of classified spectra}
I'm including some examples of various spectra. All of them are from the testing dataset and were classified using the best performing LeNet-5 architecture, \textit{l2pool}, eta decay and unbinned spectra. Red lines in pictures are absorbtion spectral lines identified by the BOSS spectroscopic pipeline (blue lines stand for emission).

\newpage
\subsection{Correctly classified objects}

\begin{figure}[!htbp]
\begin{center}
\includegraphics[width=14cm]{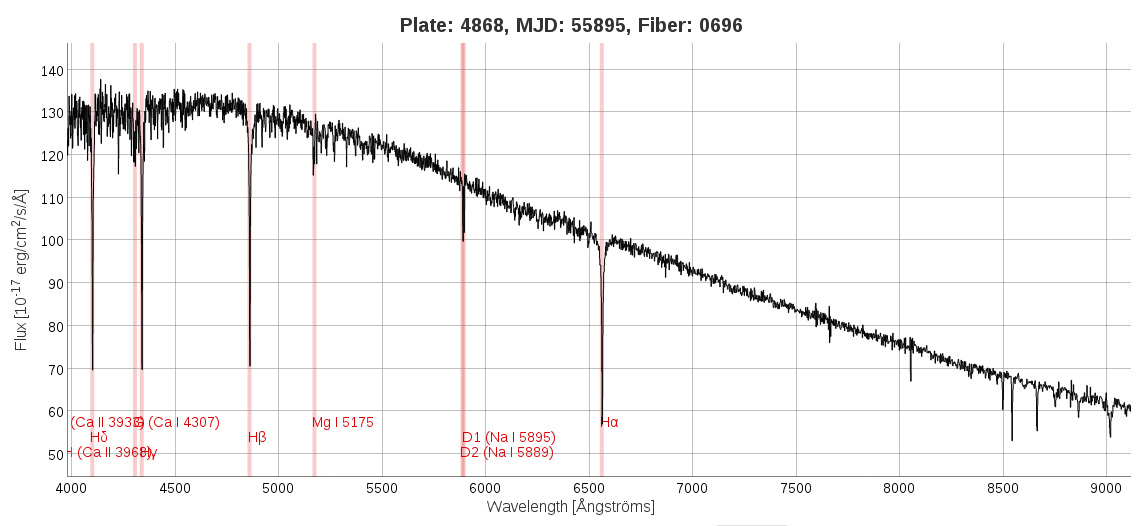}
\caption{Example of a nice stellar spectrum, absorbtion lines are clearly visible.}
\label{corr_star_nice}
\end{center}
\end{figure}

\begin{figure}[!htbp]
\begin{center}
\includegraphics[width=14cm]{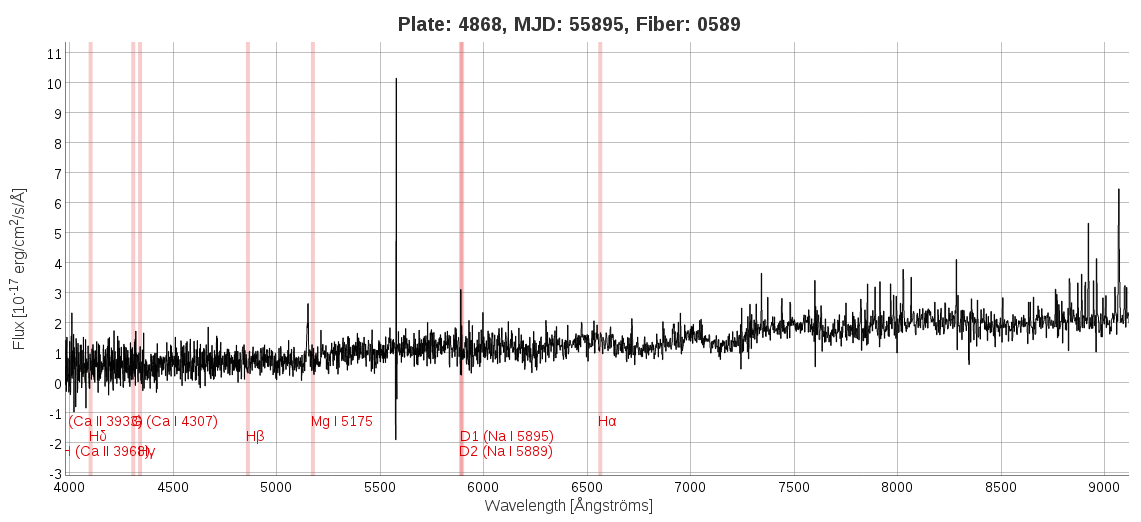}
\caption{Again a star, this time not so nice example. Still no problem for the ConvNet.}
\label{corr_star_hnus}
\end{center}
\end{figure}

\begin{figure}[!htbp]
\begin{center}
\includegraphics[width=14cm]{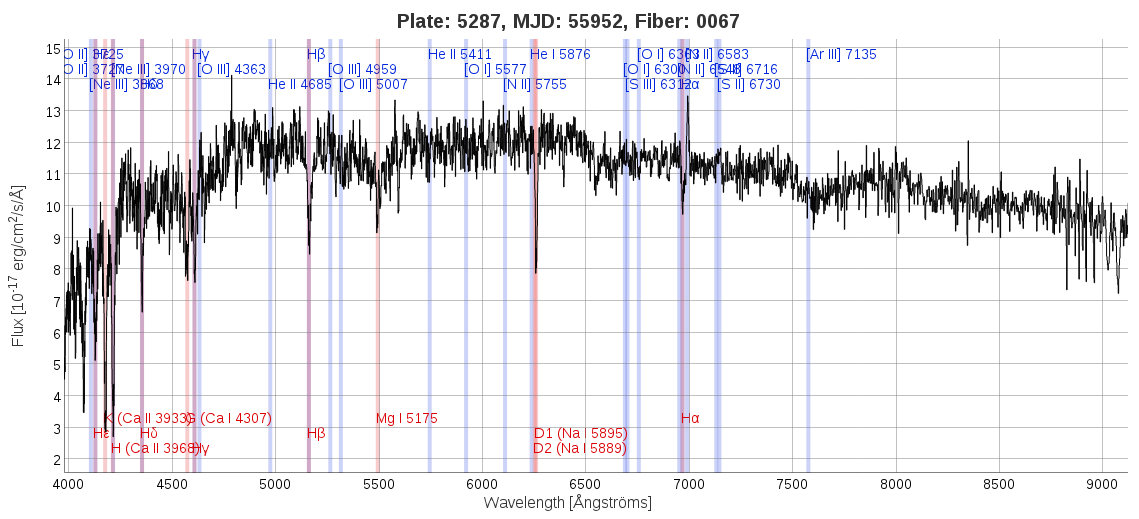}
\caption{A galactic spectrum, clearly recognizable.}
\label{corr_galaxy_nice}
\end{center}
\end{figure}

\begin{figure}[!htbp]
\begin{center}
\includegraphics[width=14cm]{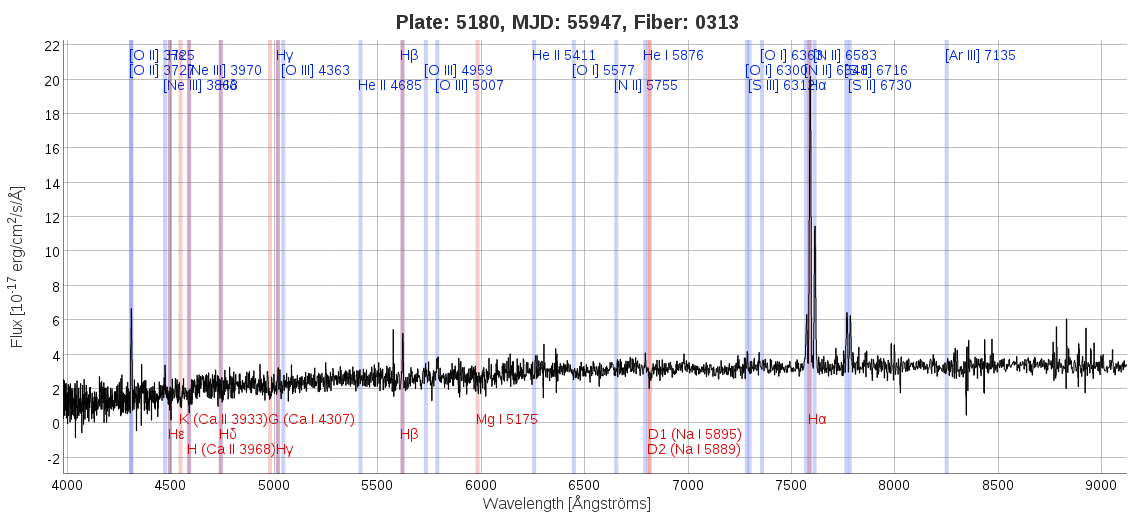}
\caption{Uncommon galactic spectrum (probabaly starforming galaxy).}
\label{corr_galaxy_hnus}
\end{center}
\end{figure}

\begin{figure}[!htbp]
\begin{center}
\includegraphics[width=14cm]{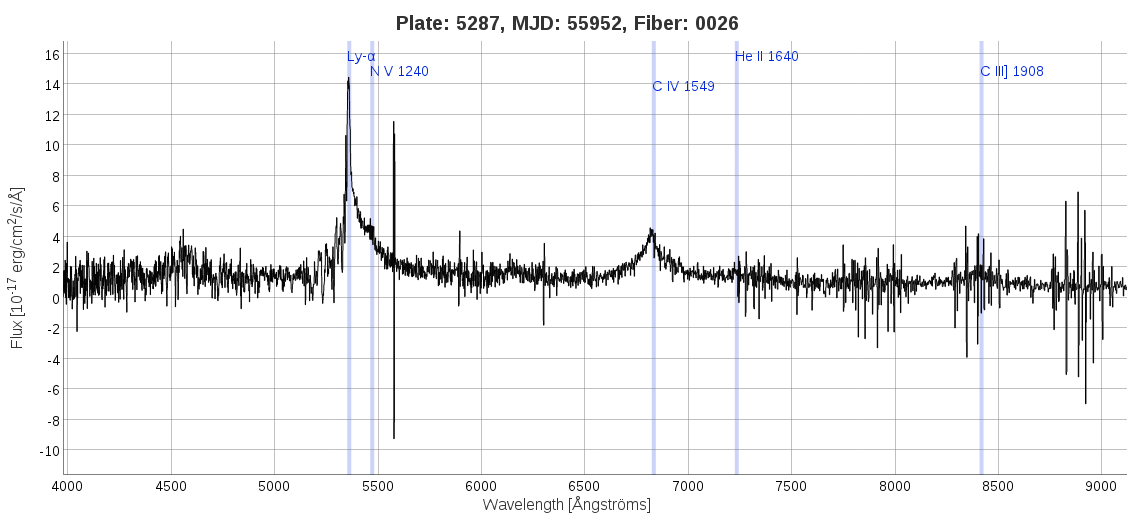}
\caption{A nice example of quasar's spectrum with broad emission lines.}
\label{corr_qso_nice}
\end{center}
\end{figure}

\begin{figure}[!htbp]
\begin{center}
\includegraphics[width=14cm]{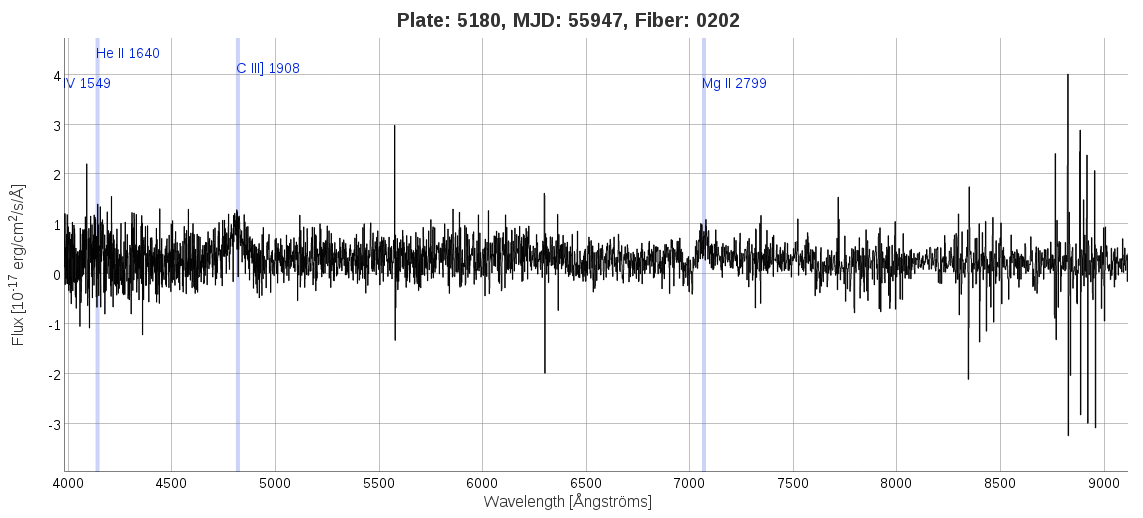}
\caption{Noisy and indistinct spectrum of a quasar.}
\label{corr_qso_hnus}
\end{center}
\end{figure}

\clearpage
\newpage
\subsection{Incorrectly classified objects}

\begin{figure}[!htbp]
\begin{center}
\includegraphics[width=14cm]{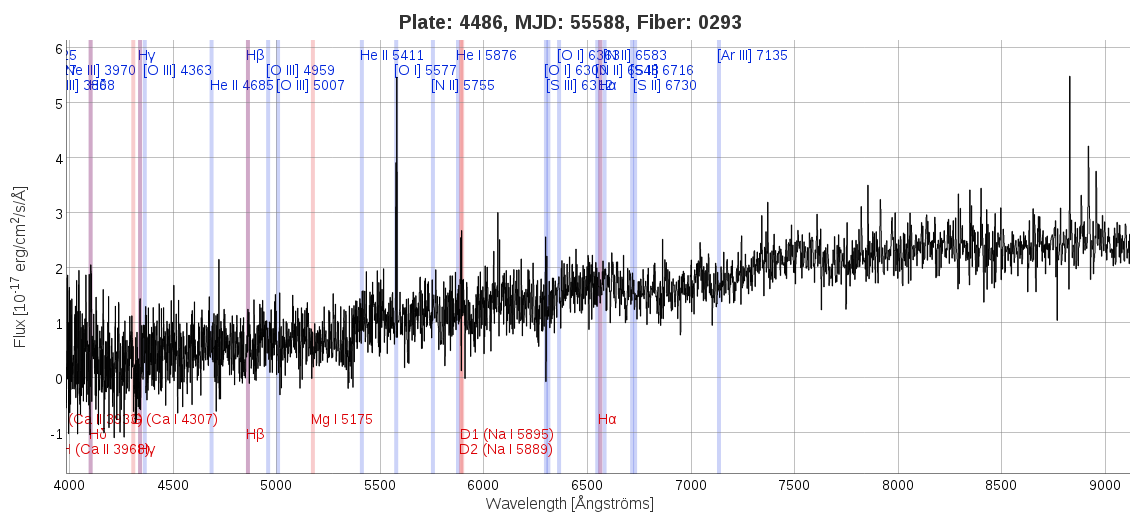}
\caption{This object is labeled as a star in the BOSS catalog. ConvNet indentified it as a galaxy.}
\label{spec_incorr_star_galaxy}
\end{center}
\end{figure}

\begin{figure}[!htbp]
\begin{center}
\includegraphics[width=14cm]{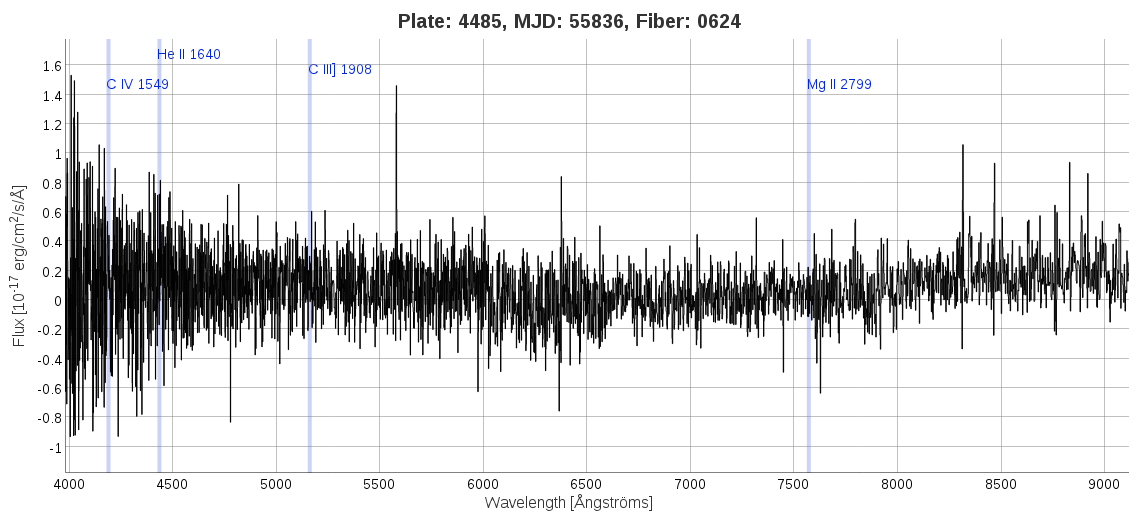}
\caption{This object is labeled as a quasar in the BOSS catalog. ConvNet indentified it as a galaxy.}
\label{spec_incorr_qso_galaxy}
\end{center}
\end{figure}

\begin{figure}[!htbp]
\begin{center}
\includegraphics[width=14cm]{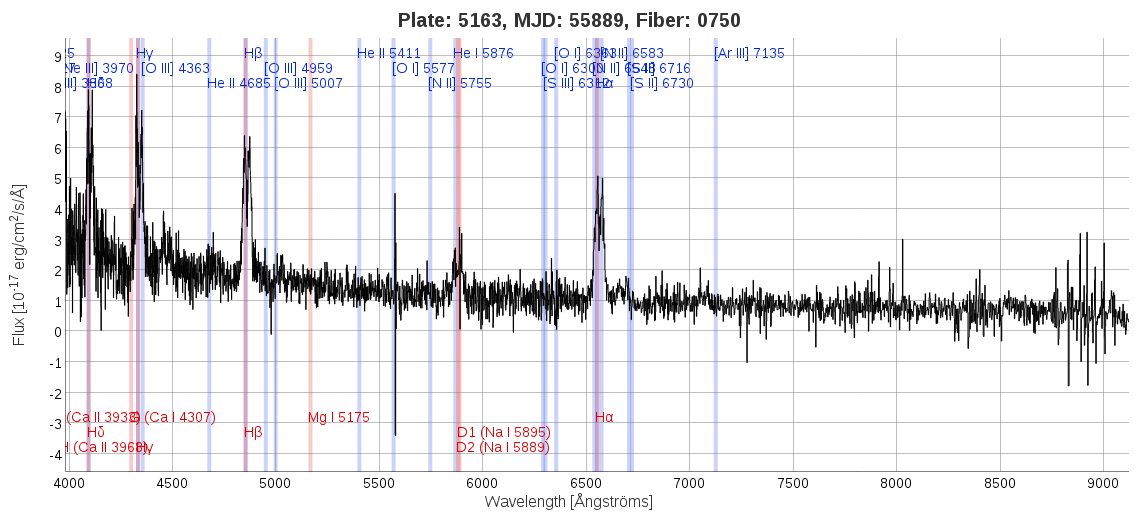}
\caption{This object is labeled as a star in the BOSS catalog. ConvNet indentified it as a quasar.}
\label{spec_incorr_star_qso}
\end{center}
\end{figure}

\subsection{Wrong labels in the BOSS catalog}

When I was browsing through the list of incorrectly classified objects in the testing dataset, I discovered there are many cases when the object's class in the BOSS catalog (fits file) doesn't match the object's class displayed on the web of the DR10 Science Archive Server. I roughly estimate maybe half of all incorrectly classified spectra are afflicted by this. When inspecting the spectra manually, I found out labels on the web are usually the correct ones. That means the ConvNet classifies the object correctly, but because the object's class in the catalog is incorrect, it's evaluated as an incorrect classification by the \texttt{classify.sh} routine. This problem artificially lowers the success rate. The best performing LeNet-5 configuration has nearly 95\% success rate on the testing dataset. Error rate is therefore 5\%. Assuming half of the incorrectly classified spectra have wrong label in the catalog, real~success rate of the ConvNet should be approximately 2.5\% higher (error rate divided by half).

But how's it even possible that there are so many incorrectly labeled spectra in the BOSS catalog? Well, maybe it isn't such a huge problem after all. My estimate that half of the incorrectly classified spectra have in fact wrong label, is only a rough estimate. I came to this number by manually inspecting few tens of objects. But there are approximately 3000 objects identified by the \texttt{classify.sh} routine as incorrectly classified. It's not humanly possible to check all these spectra. So the real number of spectra with wrong label can be lower. But even if it's the one half, that would represent about 2.5\% of all spectra, as was showed in the previous paragarph. Which is not a big deal.

What can be the real cause of this problem. Obviously I suspected the \texttt{NOQSO} to have something to do with it. So I tried using the \texttt{NOQSO} appendix for all galactic spectral parameters. But the performace of ConvNets during training and testing was then a magnitude worse (tens of \%). This was again verified by manually inspecting a subset of spectra. My decision not to use the \texttt{NOQSO} appendix is therefore right. By taking all these facts into account and considering the estimated amount of afflicted spectra to be only 2.5\%, I came to a conclusion that most probable cause of this problem is that the BOSS catalog (\texttt{specObj-BOSS-dr10.fits} file) is just some obsolete version of the true BOSS catalog. It's hard to believe this, considering the filename itself has the latest data release in its name ("dr10"). But I can't find out any other logical explanation. Clarification from somebody in the BOSS project would be very beneficial.

\newpage ~ \newpage

\chapter{Conclusion}

I have clarified the the analogy between convolutional neural networks and the human visual system. I've proved ConvNets are very powerful and efficient tool for spectral classification. The success rate of nearly 95\% on the testing dataset of more than 60000 spectra speaks by itself. Most notable fact is the simplicity and elegance of the analysis. Only spectra themselves served as an input. Preprocessing was straigtforward and the whole process required little or no human interaction. This is in clear contrast with much more complicated conventional approach of fitting wide set of (often handcrafted) spectral templates.

Considerable achievement is the software suite I've developed \cite{source}. By using the attached manual, anybody can replicate my results or use the software for his own research involving SDSS spectra. This thesis proved the artificial intelligence deserves its place in modern astrophysics. ConvNets would be most beneficial in new sky surveys, where instrumental response still carry a lot of uncertainity and conventional analysis is not very accurate (my personal experience with Fermi's LAT instrument three years ago). I can imagine trained ConvNets to be included in Virtual Oservatory, where anybody can use it to quickly determine spectral parameters of selected object even if it has never gone through regular scientific analysis.

Several interesting facts were discovered during training and classification. Contrary to expectations prior to training, simpler LeNet-5 architecture achieved better sucess rates than LeNet-7. Learning rate decay is beneficial for training. Lp-Pooling achieved better performance than subsampling in both training and testing, but its outperformance should be more deeply researched before declaring definitive winner. Surprising contrast against multilayer perceptron was the uniform performance of ConvNets. They weren't much sensitive to initial parameters. I atribute this to their deep, hierarchical structure, which enables them to abstract and brings great degree of shift and distortion invariance. Particularly important observation is the great outperformance of configurations using unbinned spectra. This implies the spectral lines are much more important features for spectral classification than shape of the continuum.

If not for the wrong labels in the BOSS catalog, average success rates on testing dataset may be as much as 2\% higher. And by looking at the spectra that were truly incorrectly classified, they are often very noisy and hard to classify even for an experienced astrophysicist. It would be therefore very interesting to compare my results with real humans. It's probably a strong statement, but if the quality of the training dataset is further improved, I think it won't be unimaginable for my algorithm to achive better success rate in spectral classification than is humanly possible. It wouldn't be the first time the convolutional neural network beat the humans \cite{ach_traffic}.

\appendix

\chapter{Software manual}
This is a manual for the software pack composed of bash scripts and C programs I developed for the purpose of this thesis. Source codes are available at this link: \cite{source}. This step by step manual will guide you through datasets preparation, preprocessing of spectra and ConNet's construction/training/classification. All the source codes are attached to this thesis \cite{source}, C programs have to be compiled. Beside them, you will need the EBLearn library \cite{eblearn}, MAT File I/O Library, heatools package from the HEAsoft pack \cite{heasoft}, GNU coreutils, imagemagick utils, awk and gnuplot. The software was ran and debugged on amd64 Debian distribution of GNU/Linux.

All bash scripts and compiled C programs should be placed into a three-folder structure:

\begin{center}
\includegraphics[width=10cm]{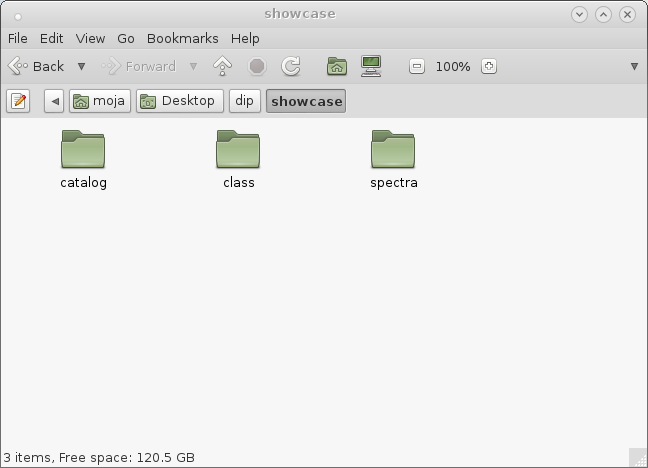}
\end{center}

Datasets preparation will happen in the \textit{catalog} folder, spectral preprocessing in the \textit{spectra} folder and ConvNet handling in the \textit{class} folder.

\section{Datasets preparation}
Let's begin with the \textit{catalog} folder. These files should be located in this folder by default:

\begin{center}
\includegraphics[width=10cm]{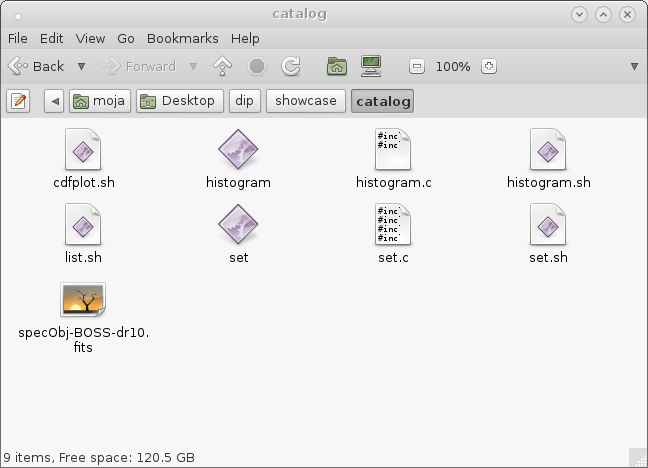}
\end{center}

The fits file is the BOSS catalog. \texttt{list.sh} is the first script we have to run. Make it executable and don't forget to initialize heatools first:

\begin{lstlisting}
shell$ heainit
shell$ ./list.sh 
\end{lstlisting}

The script uses heatools to extract required spectral features and identifiers into text files and reduces those lists, so they can be processed by other routines. Next step is creation of training, validation and testing datasets. This is done by \texttt{set.sh} script and \texttt{set} program compiled from \texttt{set.c} source code. Open the \texttt{set.sh} file first and look for these lines:

\begin{lstlisting}
	if [[ $file == *qso ]]; then size_of_the_set="16850"; fi
	if [[ $file == *galaxy ]]; then size_of_the_set="11680"; fi
	if [[ $file == *star ]]; then size_of_the_set="34200"; fi
\end{lstlisting} 

This block of lines is with slight variations presented on three places in the code. First for the training set, second for the validation set and third for the testing set. Size of subsets for each dataset depends on the green parameters (variable \texttt{size\_of\_the\_set}). By adjusting these numbers, you influence what will be the size of each subset (galaxy, qso, star) in each dataset (train, valid, test). But remember they doesn't correspond to actual size of each subset. Numbers set in the script by default correspond to datasets sizes and distributions as specified in table \ref{tab_sets}. Then simply run the \texttt{set.sh} script (don't forget to make it executable):

\begin{lstlisting}
shell$ ./set.sh 
\end{lstlisting}

Datasets created by this script are saved into the \textit{sets} folder. \textit{catalog} folder then looks like this:

\begin{center}
\includegraphics[width=10cm]{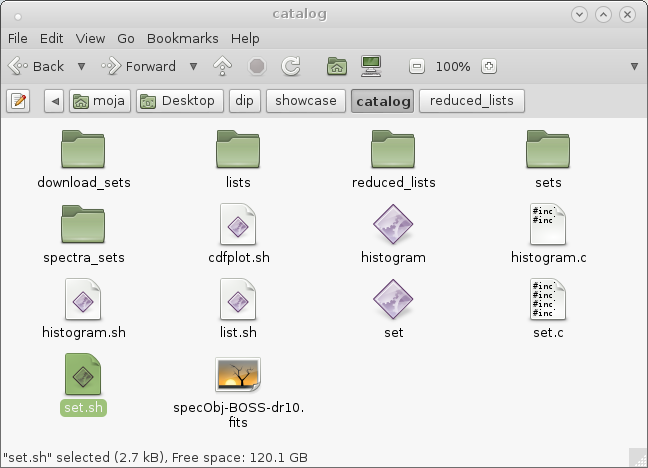}
\end{center}

The \textit{sets} folder itself contains these text files:

\begin{center}
\includegraphics[width=10cm]{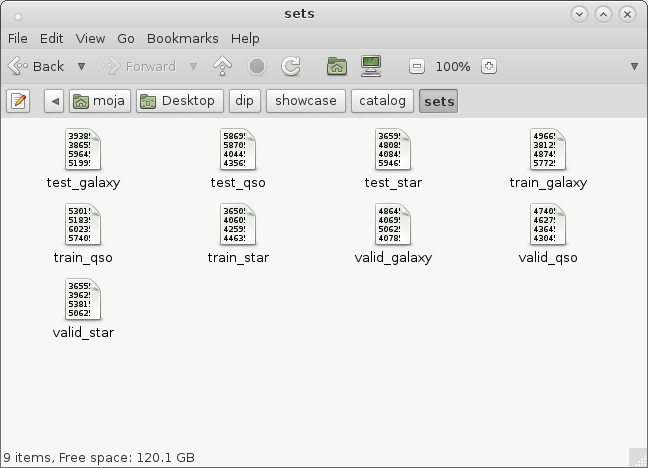}
\end{center}

First part of the file's name specifies the dataset (train, valid, test) and second part the subset (object class). However the \textit{set} folder isn't the important one. It serves as a source of data for generating histograms and cumulative distribution function charts (using \texttt{histogram.sh} script or \texttt{cdfplot.sh} for comparative studies), but nothing more. The most important folder is the \textit{spectra\_sets}. It contains lists of objects in each dataset.

\begin{center}
\includegraphics[width=10cm]{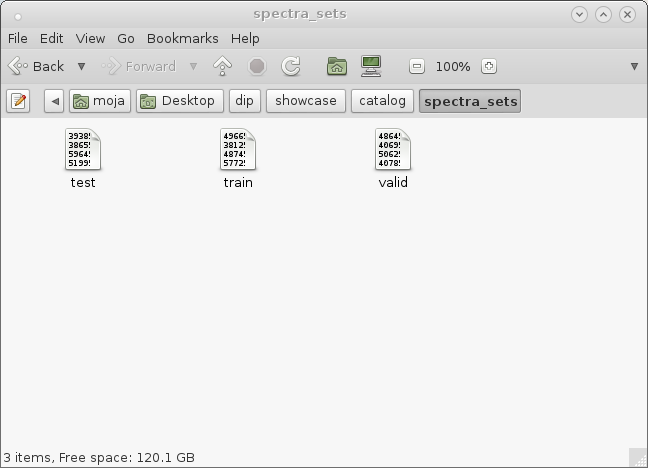}
\end{center}

For instance here is an excerpt of few lines from the \texttt{train} file with columns description:

\begin{lstlisting}
#PLATE	MJD	 FIBERID CLASS	REDSHIFT
5169	56045	524	1	3.5992600000
3819	55540	279	1	3.5994100000
4210	55444	184	1	3.6002400000
5890	56037	504	1	3.6002500000
5484	56039	109	1	3.6008700000
3770	55234	160	1	3.6011600000
5992	56066	454	1	3.6015000000
5459	56035	554	1	3.6018900000
...		...		...	...	...
\end{lstlisting}

The CLASS label is encoded (0=galaxy, 1=qso, 2=star). \textit{spectra\_sets} will be later used for spectral preprocessing and classification. \textit{download\_sets} folder is similar to \textit{spectra\_sets}, but text files in it contains only identifiers (PLATE, MJD, FIBERID). These lists will be used to download spectra from the SDSS Science Archive Server.

\newpage
\section{Preprocessing}
Preprocessing of spectra will take place in the \textit{spectra} folder. This is the folder's structure by default:

\begin{center}
\includegraphics[width=10cm]{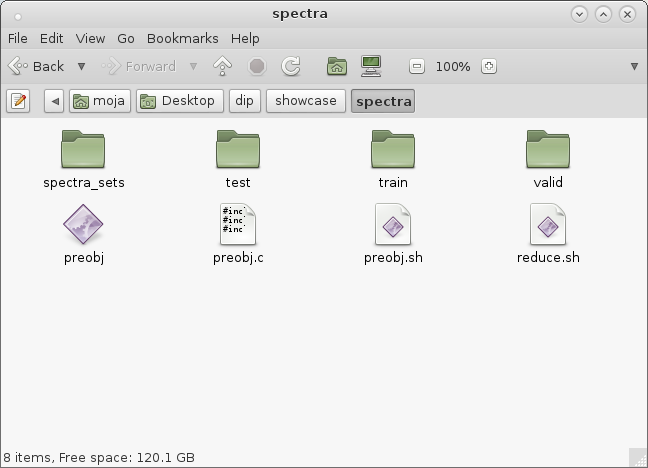}
\end{center}

Folder \textit{spectra\_sets} is just copied from the \textit{catalog} folder. \textit{train}, \textit{valid} and \textit{test} folders contain spectra (fits files) downloaded from the Science Archive Server. Only thing you have to do is to simply run the \texttt{reduce.sh} script (if you closed the terminal after the preparation of datesets, reinitalize heatools):

\begin{lstlisting}
shell$ ./reduce.sh 
\end{lstlisting}

This operation extracts the spectra from fits files, reduces them and does several other modifications. There is one important thing to remember. This operation alters the \textit{spectra\_sets} folder (removes blacklisted spectra from datasets). Therefore if you wish to run \texttt{reduce.sh} script again (for instance trying different parameters), don't forget to replace the altered \textit{spectra\_sets} folder with the original one from \textit{catalog} folder.

Next step is the \texttt{preobj.sh} script. It bins spectra and transforms them from one dimensional vectors into two dimensional matrices (and then converts them into images). Find out following lines in the script:

\begin{lstlisting}
	mati="60"
	matj="60"
\end{lstlisting}

They specify dimensions of the output matrix. You can adjust these dimensions, but keep two things in mind. The output matrix has to be a square and number of its elements has to be smaller or equal than number of elements in the input vector. Then you can simply execute the script:

\begin{lstlisting}
shell$ ./preobj.sh 
\end{lstlisting}

The folder at the end contains these files:

\begin{center}
\includegraphics[width=10cm]{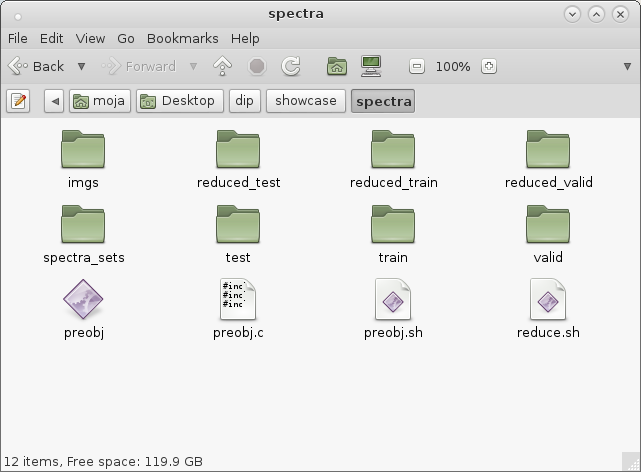}
\end{center}

\textit{imgs} folder contains spectral images sorted into directories by dataset and subdirectories by object class. Content of one of these subdirectories (only fraction) can look like this:

\begin{center}
\includegraphics[width=10cm]{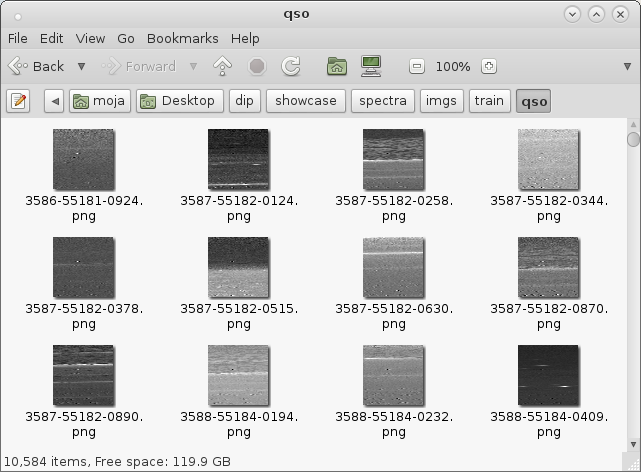}
\end{center}

\section{Training and classification}

I will describe a sample training of the convolutional neural network and classification on the testing dataset. Our working environment will be the \textit{class} folder:

\begin{center}
\includegraphics[width=10cm]{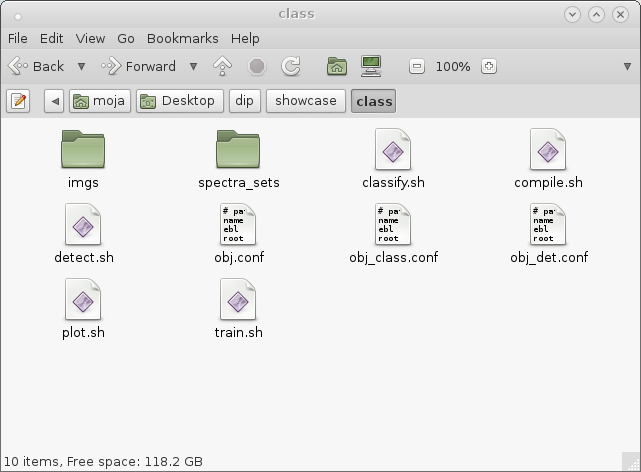}
\end{center}

\textit{imgs} and \textit{spectra\_sets} directories were copied from the \textit{spectra} folder. Files with \texttt{.conf} extension are configuration files for EBLearn library. I won't go into much detail about their creation and modification because it would be beyond this manual. You can use files attached in the Appendix B, they are configuration files for the best performing LeNet-5 architecture with \textit{l2pool} and eta decay. These attached configuration files will work. You only have to adjust paths:

\begin{lstlisting}
# paths ###################################################################
name            = object
ebl             = /home/moja/eblearn_1.2_r2631          #eblearn root
root            = /home/moja/Desktop/dip/data/class/lenet5_pool_decay_i100/inputs		#data root
tblroot         = ${ebl}/tools/data/tables # location of table files
\end{lstlisting}

\texttt{ebl} variable is path to the EBLearn library and \texttt{root} is path to the compiled inputs directory (product of the \texttt{compile.sh} script). To train the network, you have to run \texttt{compile.sh} and then \texttt{train.sh} scripts. Only parameter you have to adjust is the

\begin{lstlisting}
dim="60"
\end{lstlisting}

parameter in \texttt{dscompile.sh} script (it's also in \texttt{dscompile.sh}). It's the dimension of input images. Then execute the scripts:

\begin{lstlisting}
shell$ ./compile.sh
shell$ ./train.sh 
\end{lstlisting}

Product of training is the \texttt{train\_output} file, which sums up the training process and \textit{net} directory that containes trained network configuration files for each epoch. You can visualize the training process (success rates) by running the \texttt{plot.sh} script. Classification on the testing dataset is performed by \texttt{classify.sh} script:

\begin{lstlisting}
shell$ ./classify.sh
\end{lstlisting}

The \textit{class} folder will have this structure at the end:

\begin{center}
\includegraphics[width=10cm]{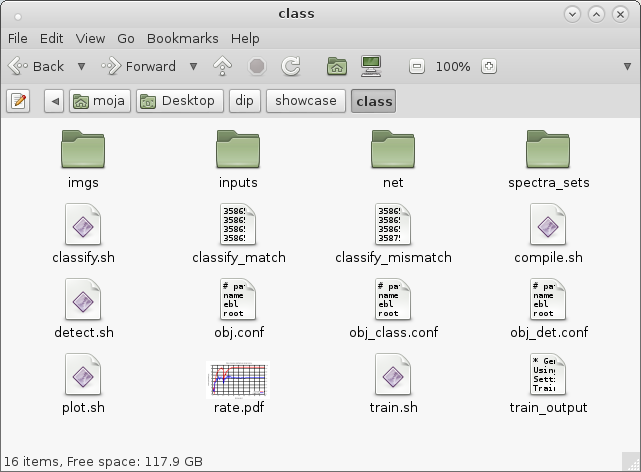}
\end{center}

\texttt{classify\_match} is a product of the classification. It contains correctly classified objects from the testing dataset (with labels). \texttt{classify\_mismatch} on the other hand contains objects where ConvNet's classification is in conflict with the BOSS catalog. You can compute success rate on the testing dataset by comparing these two files. 

Here's an excerpt from the \texttt{classify\_mismatch} file:

\begin{lstlisting}
#PLATE	MJD		FIBERID
3586	55181	45		catalog: qso		convnet: galaxy
3586	55181	94		catalog: galaxy		convnet: star
3587	55182	733		catalog: star		convnet: qso
3587	55182	739		catalog: galaxy		convnet: qso
3588	55184	70		catalog: qso		convnet: galaxy
3588	55184	280		catalog: qso		convnet: galaxy
...		...		...		...					...
\end{lstlisting}

\bibliographystyle{plain}
\bibliography{literatura}

\end{document}